\documentclass[final,5p,times,twocolumn,authoryear,longnamesfirst]{elsarticle}

\usepackage{graphicx}
\usepackage{amssymb}
\usepackage{xcolor}
\usepackage{subcaption}
\usepackage{hyperref}

\begin{document}

\begin{frontmatter}

\title{Longitudinal Abuse and Sentiment Analysis of Hollywood Movie Dialogues using Language Models}

\author[inst1,inst2]{Rohitash Chandra\corref{cor1}}  
\ead{rohitash.chandra@unsw.edu.au}

\author[inst1,inst2]{Guoxiang Ren}
\ead{guoxiang.ren@unsw.edu.au}  

\cortext[cor1]{Corresponding author: Rohitash Chandra (rohitash.chandra@unsw.edu.au)}

\address[inst1]{Transitional Artificial Intelligence Research Group, School of Mathematics and Statistics, UNSW Sydney, Sydney, Australia}
\address[inst2]{Centre for Artificial Intelligence and Innovation, Pingla Institute, Sydney, Australia}




\begin{abstract}
Over the past decades, there has been an increase in the prevalence of abusive and violent content in Hollywood movies. In this study, we use \textcolor{black}{language models} to explore the longitudinal abuse and sentiment analysis of Hollywood Oscar and blockbuster movie dialogues from 1950 to 2024. \textcolor{black}{We provide an analysis of subtitles for over a thousand movies, which are categorised into four genres. We employ fine-tuned language models to examine the trends and shifts in emotional and abusive content over the past seven decades.} Findings reveal significant temporal changes in movie dialogues, which reflect broader social and cultural influences. Overall, the emotional tendencies in the films are diverse, and the detection of abusive content also exhibits significant fluctuations. The results show a gradual rise in abusive content in recent decades, reflecting social norms and regulatory policy changes. Genres such as thrillers still present a higher frequency of abusive content that emphasises the ongoing narrative role of violence and conflict. At the same time, underlying positive emotions such as humour and optimism remain prevalent in most of the movies. Furthermore, the gradual increase of abusive content in movie dialogues has been significant over the last two decades, where Oscar-nominated movies overtook the top ten blockbusters.
\end{abstract}


\end{frontmatter}

\section{Introduction}
\label{sec:sample1}

The cinema has been present in various forms for over a century, serving multiple purposes as an artistic medium through movies and series \citep{nowell1996oxford}. Since the age of television \citep{buonanno2008age}, the cinema has evolved from a mere source of entertainment to an influential tool that includes politics \citep{tzioumakis2016routledge}, fashion, and cultural and social changes \citep{kerrigan2018movies}, supporting activism and policy changes, e.g., advocacy for queer activism and same-sex marriage \citep{schoonover2015worlds}. Furthermore, there has also been activism with movies about climate change \citep{sakellari2015cinematic} including natural disasters (e.g. Day After Tomorrow \citep{leiserowitz2004day}) and dystopian futures showing a water crisis (e.g. Mad Max \citep{hassler2017mad}). Although cinema has played an important role in shaping culture and inspiring activism, it has also been criticised for its potential negative influences, particularly through the depiction of violence and abusive content.

The cinema had profound positive and negative cultural and social effects. Some of the negative impacts are an increase in crime rate \citep{young2009scene}, such as drug use and distribution \citep{grist2013drugs}, and sex crimes \citep{gunasekera2005sex}.
Violence is a powerful narrative tool in cinema, despite its adverse social effects.  Violence and abuse in cinema have been used to enhance a viewer’s experience, and boost movie profits \citep{barranco2020ticket}; therefore, the prevalence of violence in movies has significantly escalated over the past few decades \citep{markey2015violent}. However, the relationship between movies and human behaviour is of utmost significance \citep{fearing1947influence} and violent movies often have an inciting effect \citep{Dunand1984}. Violent movies promote aggression and lead to antisocial behaviour \citep{berkowitz1963film} that includes aggressive behaviour as well as cognitive, social, and emotional problems, often accompanied by a lack of empathy \citep{paik1994effects}.   Anderson and Bushman \citep{anderson2002effects}  presented a meta-analysis and reported strong links between exposure to violent media and increased aggression in viewers. \textcolor{black}{Later, they provided a theoretical synthesis explaining how exposure to violent media affects aggression using three pathways: cognitive, affective, and arousal mechanisms  \citep{anderson2018media}.} The cinema includes movies, series, and documentaries, which can have lasting emotional effects on young viewers. Canter et al.  \citep{cantor2001media} reported that different people have different emotional and behavioural reactions to violent movies. Moreover, they reported that children's exposure to cinema and related media can lead to increased fear and anxiety. \textcolor{black}{These social concerns are further amplified in the contemporary media landscape, where digital platforms have transformed how audiences consume cinematic content.}

The way of obtaining entertainment has become more diverse with the Internet and emerging technologies. OTT (Over-the-top) \citep{varghese2021ott} is a digital media platform for film and television content delivered through the Internet. OTT service providers such as Netflix, Amazon Prime Video and Disney+ use a subscription system or pay-per-view to provide users with a huge amount of movie and television resources, which has become an important channel for current entertainment consumption \citep{eklund2023streaming}.
The future could have OTT platforms gaining the upper hand over traditional movie theatres due to factors, such as diverse content, language availability, ease of viewing, and affordable costs \citep{gaonkar2022ott}. As OTT platforms become one of the main channels for entertainment, \textcolor{black}{The study will} also examine the challenges that may arise. Given the scale and diversity of content available on these platforms, computational approaches such as Natural Language Processing (NLP) have become essential for systematically analysing large volumes of textual data, including film dialogues.
The field of  NLP \citep{cambria2014jumping} lies at the confluence of artificial intelligence and linguistics, involving the design and implementation of algorithms that interact with human language using methods such as deep learning \citep{9075398}. Over the past few years, the widespread use of deep learning models has propelled NLP, enabling it to deal efficiently with more complex language problems \citep{9075398}. Sentiment analysis \citep{medhat2014sentiment} using deep learning models \citep{ain2017sentiment}, focuses on detecting human emotions in text, such as 'anxious', 'happy', 'optimistic' and sentiment polarity prediction, such as positive, negative, or neutral opinions. Sentiment analysis has extensive applications in different domains, encompassing social media analytics, public sentiment monitoring, product review evaluation, and customer feedback assessment \citep{dang2020sentiment,medhat2014sentiment}. Recurrent neural networks (RNNs) \citep{salehinejad2017recent} are deep learning models that possess the ability to acquire features and capture long-term dependencies from sequential and time-series data \citep{salehinejad2017recent}. The Transformer model is a deep learning model that features an encoder-decoder architecture and attention-based mechanism inspired by cognitive science  \citep{vaswani2017attention}. Although RNNs have been widely used, the introduction of the Transformer model has marked a significant shift in NLP, paving the way for pre-trained \textcolor{black}{language models}.
Recent innovations in pre-trained language models, also known as \textcolor{black}{language models} \citep{kasneci2023chatgpt}, include BERT (Bidirectional Encoder Representation from Transformers) \citep{devlin2018bert} and GPT (Generative Pre-trained Transformers) \citep{kasneci2023chatgpt}, which have become prominent in a wide range of NLP tasks, exemplified by implementations like GPT-4o, Gemini 2.5, and Grok 3.

Abuse detection using language models has been prominent due to the needs of social media, where real-time management of discussions are needed to avoid cyber-bullying. Caselli et al. \citep{caselli2020hatebert} introduced the Hate-BERT model for abusive content detection, achieving better accuracy in detecting various abusive content categories including offensive content, abusive content, and hateful language.  Nobata et al. \citep{nobata2016abusive} effectively developed a machine learning-based approach to detect hate speech in online user comments.  Mozafari et al. \citep{mozafari2020bert} added a Bidirectional-LSTM layer to BERT's architecture to enhance its classification capabilities. Mnassri et al. \citep{mnassri2022bert} explored various integrated approaches to improve BERT for hate speech detection, and reported that combining BERT with other deep learning models improved accuracy. Although these studies demonstrate the effectiveness of language models for abuse detection in social media contexts, their application to long-term cinematic data remains underexplored.

The defining characteristic of longitudinal analysis lies in the repeated measurements taken on the same individuals or subjects \textcolor{black}{of} study over time, enabling a direct examination of temporal changes \citep{fitzmaurice2012applied}. The primary objective of a longitudinal study is to delineate the dynamics of response variation and identify influential factors \citep{fitzmaurice2012applied}. One can effectively examine temporal variations \citep{fitzmaurice2012applied} by capturing within-individual changes through multiple assessments. \textcolor{black}{This motivates our study, which combines abuse detection and sentiment analysis to examine temporal trends in cinematic dialogues over more than seventy years.}

In this study, \textcolor{black}{we present a framework} for abuse detection and sentiment analysis on dialogues from selected movies. \textcolor{black}{Hollywood movies (films) spanning from 1950 to 2024 were selected} based on Oscar nominations and box-office collections. \textcolor{black}{We analyse} decadal changes in the levels of abuse and violence, utilising sentiment analysis across major categories, including Oscar nominations and blockbuster hits. We classify the selected movies into four categories to provide an analysis of how the abuse and sentiment analysis impact different categories of movies, e.g. action vs drama. An abuse detection dataset \citep{caselli2020hatebert} is used to fine-tune. BERT-based models and provide a longitudinal study. The framework also provides sentiment analysis, with the language models fine-tuned using the SenWave \citep{yang2020senwave} dataset. The subtitle dataset of more than a thousand movies was also curated for further analysis. 

This study addresses several gaps in the existing literature. First, prior studies on sentiment and abuse analysis have primarily focused on social media or reviews, with limited longitudinal analysis of cinematic dialogues across decades. Second, most studies focus on a single dimension (either sentiment or abusive content),  \textcolor{black}{this work integrates} both to provide a more holistic perspective. Third, previous studies have not compared Oscar-nominated and blockbuster films and examined cross-genre differences. The contributions of this study are threefold:

\textcolor{black}{(i) Curation and analysis of a large-scale subtitle corpus covering more than seven decades of Hollywood films.}

\textcolor{black}{(ii) Fine-tuning of advanced language models for simultaneous sentiment and abuse detection.}

\textcolor{black}{(iii) Provision of comparative insights across genres and award/commercial classifications, highlighting temporal shifts in cinematic language use.}

 The structure of the paper is as follows: Section 2 provides background information, and Section 3 discusses the methodology, including the framework, data processing, and modelling. In Section 4, the results are presented. Section 5 offers a discussion and outlines directions for further research. Finally, Section 6 concludes the study.

\section{Background}
\subsection{Cinema: impact on society}

Film as a medium is not only considered one of the most important sources of entertainment, but also a powerful tool for influencing society and culture \citep{saeed2018impact}.
Although the themes of love, friendship, and family relationships in films inspire audiences positively, some films with explicit, vulgar, or violent content can negatively impact viewers’ psychological well-being \citep{worth2008exposure}, affecting children and adolescents \citep{agarwal2012harmful}
 . Furthermore, the characteristics of the protagonist, such as justice, wisdom and bravery, have a positive influence on the growth of young people \citep{smithikrai2016effectiveness}.  
 
The United Nations Educational, Scientific and Cultural Organisation (UNESCO) \citep{groebel1998unesco} presented a comprehensive study of the impact of media violence on the audience. Most studies show that the relationship between media abuse and “real” abuse is interactive, and the perpetuation and persistence of media violence fuelling a global culture of aggression \citep{ybarra2008linkages,morahan1999relationship,mishna2009real}. An empirical study \citep{ferguson2015does} in 2015 challenged simplistic causal assumptions, showing that the relationship between media violence and societal aggression is highly context-dependent and varies across time periods. The American psychologist, Albert Bandura \citep{Bandura1977social} reported that when people consciously or unconsciously remember and preserve abusive content, and unconsciously activate it, there are setbacks in social interactions. According to the American Academy of Pediatrics \citep{10.1542/peds.2009-2146}, television violence accounts for 10 percent of the sources of learning about violent behaviour in children. Vitasari \citep{vitasari2013analysis} reported that the use of abusive words is now more often used in films. \textcolor{black}{Using a corpus of film dialogues, Pavesi and Formentelli \citep{pavesi2023pragmatic} found that swearing serves coherence and emotional emphasis in audiovisual translation, rather than mere profanity.} The uninhibited display of emotions is an indication of cultural transformation, resulting in a rise in the utilisation of profane language. Hasibuan \citep{hasibuan2021analysis} contended that many films now often use swearing in dialogue to make them more interesting. Novari \citep{novari2022usage} reported that characters in movies do not always use swearing to curse but to express their other emotions and feelings in different situations. An editorial synthesis \citep{stapleton2023swearing} argued that swearing functions as a pragmatic strategy to convey interpersonal nuance beyond explicit meaning.  Baid et al. \citep{baid2017sentiment} demonstrated that morphing learning from the social media domain can effectively classify hateful and offensive speech in movie subtitles. The authors used domain-adaptive and fine-tuned transfer learning techniques on an existing social media dataset. Deep learning models have been prominent in sentiment analysis \citep{chandra2023analysis}; hence, \textcolor{black}{the study} use them to analyse movie dialogues.

\subsection{Web series and movies}

 Web series \citep{christian2018open} is a short-form interstitial content distributed through online platforms.
 Since the 1990s, with the advent of technological change and the emergence of the new media market, web series have become prominent, splitting the market from limited television and film \citep{christian2012web}. During COVID-19, \textcolor{black}{as countries imposed lockdowns and economic activity slowed, online sketches gained a significant opportunity to grow and reach more people.} Traditional movies have been facing a huge challenge due to OTT platforms, and a study shows that as the length of movies increases, people expect movies released in theatres to be released simultaneously on OTT as well \citep{yan2021relationship}. Therefore, web series have become one of the most popular trends on OTT platforms \citep{deb2022proliferation}. However, the content preferences for films and web series are very different.
In a related study, Wadhwa et al. \citep{wadhwa2020new} conducted sentiment analysis for ten films of different genres as well as ten web series during 2017-2019. The authors accessed Twitter data related to each show through hashtags, and 3000 tweets were counted to determine the sentiment of each web series and categorised into positive and negative sentiment. The results found that horror web series were more popular with viewers than horror films, and categories such as comedy, romance and crime films were more popular than web series.


\section{Methodology}

\subsection{Data}

The Oscars (Academy Awards) \footnote{Oscar website: \url{https://www.oscars.org/}}\textcolor{black}{is presented annually by the Academy of Motion Picture Arts and Sciences (AMPAS) and primarily recognises artistic and technical achievements in Hollywood movies} \citep{OscarMovies}. Historically, the Oscars have nominated a wide range of films, including historical dramas, romances, and science fiction. Although the drama category \textcolor{black}{has traditionally dominated} the Academy Awards, horror, comedy, and independent films have gradually gained more recognition in recent years with movies such as "Get Out" and "Backstage." This diversity allows Oscar-nominated films to represent different periods and audiences' aesthetic tastes \citep{OscarMovies}. The annual top 10 global box office films represent the most popular movie genres and themes. These films typically attract large audiences and exert wide global influence. They span genres such as action, science fiction, comedy, and animation. Their global reception offers insights into cultural preferences and emotional resonance. The popularity of these films in different cultural contexts reflects the general acceptance of their emotional expressions and scenes of violence. Therefore, we focused on the subtitles of Oscar-nominated movies and the top 10 movie subtitles at the box office each year from 1950 to 2024.

We obtained the top 10 Oscar-nominated movies and top 10 movies at the global Box Office for 1950 to 2024  from a website known as boxofficemojo \footnote{\url{https://www.boxofficemojo.com/}}. \textcolor{black}{We extracted the subtitles (dialogues) for the top 10 Oscar-nominated and global Box Office films from 1950 to 2024 from Opensubtitles \footnote{\url{https://www.opensubtitles.com/en}}}. \textcolor{black}{The number of Oscar-nominated films varies across years. In total, we collected 1026 film subtitles, grouped by year and genre to analyse temporal trends in violence, abusive content, and emotional expression in films.} The dataset is available via Kaggle \footnote{\url{https://www.kaggle.com/datasets/mlopssss/subtitles}} and also documented in our GitHub repository \footnote{\url{https://github.com/sydney-machine-learning/sentimentanalysis-Hollywood}\label{github}} for this study.

We classify the movies into four broad groups based on IMDb (International Movie Database) \footnote{\url{https://www.imdb.com/}}
definitions, merging similar genres for comparison. We merged action, adventure, and crime into a single category called "Action". We combineed Horror and thriller into the "Thriller" category. We keep the "Comedy" and "Drama" categories unchanged. Since most movies are defined by three category labels on the IMBb website, we take the first label as the main category of the movie as the basis for classification and reports the number of movies under each category in Table \ref{tab:filmcategories}. 

In addition to movie subtitles, we incorporated external datasets to train sentiment and abuse detection models. The SenWave \citep{yang2020senwave} dataset for sentiment analysis comprises over 105 million collected tweets worldwide, providing an assessment of global sentiment fluctuations during the COVID-19 pandemic. This comprehensive dataset includes annotations for 10,000 English and 10,000 Arabic tweets across ten distinct categories: optimism, gratitude, compassion, pessimism, anxiety, sadness, anger, denial, official reports, and jokes. Qiang et al. \citep{loper2002nltk} preprocessed the raw data by filtering irrelevant characters using the NLTK (Natural Language Toolkit) tool.  The SenWave dataset has been used to identify post-COVID-19 sentiment trends \citep{chandra2021covid} and to refine pre-trained language models for machine translation in low-resource languages \citep{chandra2022semantic}. These studies motivate the usage of the SenWave data for the analysis of Hollywood movie dialogues. 

RAL-E (Reddit Abusive Language English) \citep{caselli2020hatebert} is a large-scale dataset \textcolor{black}{comprising} 1,478,348 messages from the Reddit \footnote{\url{www.blackdit.com}} online communities called \textit{Subredits} \citep{ballinger2022using} that were banned for offensive, hateful, and abusive content. Subreddits were created and controlled by users, with moderators setting community rules \citep{matias2016going}. In June 2015, Reddit banned hate speech-related Subreddits for violating the site's \textcolor{black}{user agreement and updated its content policy \footnote{\url{https://www.redditinc.com/policies/content-policy}}.} Tommaso et al. \citep{caselli2020hatebert} analysed the abusive vocabulary and compiled a data set for further research. The RAL-E data set has been used to train several transformer-based deep learning models, which have shown impressive performance in detecting online abuse \citep{caselli2020hatebert,huang2022multitask,dacon2022detecting}.
This study uses the RAL-E dataset to fine-tune the HateBERT model and improve its abuse detection capabilities.

\begin{table}[h!]
\centering
\begin{tabular}{|c|c|}
\hline
\textbf{Categories} & \textbf{Number} \\
\hline
Action & 478 \\
\hline
Comedy & 222 \\
\hline
Drama & 285 \\
\hline
Thriller & 41 \\
\hline
\end{tabular}
\caption{Number of movies in the four categories that include drama, action, comedy and thriller.}
\label{tab:filmcategories}
\end{table}


\subsection{N-gram analysis}

The N-gram model is a basic NLP method used to examine 'n' consecutive words from a given sequence of text \citep{tripathy2016classification}. The n-gram model has been widely used in text categorisation and provides additional information about sentiment classification \citep{dey2018senti}  and improves the classification performance of informal text-based tasks \citep{kruczek2020n}. The simplicity and efficiency of the n-gram model make it the preferred choice for large vocabulary speech recognition applications \citep{liu2003use}. Pang et al. \citep{pang2002thumbs} employed unigram, bigram, and a combination of both unigram and bigram techniques to classify positive and negative sentiments for classification. In the framework, we used bigrams and trigrams to ascertain the frequency of word occurrences, combined with sentiment analysis.

\subsection{Model Selection and Fine-tuning Strategy}

We selected RoBERTa (Robustly Optimised BERT) \citep{liu2019roberta,zhu2020identifying} as it features an enhanced training approach with larger corpora, longer text sequences, and dynamic masking patterns, which proved more effective than standard BERT when fine-tuned with the SenWave dataset.

We employed HateBERT \citep{caselli2020hatebert} for abuse detection, a BERT variant specifically designed for hate speech detection. HateBERT's pre-training on abusive content from Reddit communities makes it particularly suitable for our movie dialogue analysis task. Our fine-tuning approach involves domain-specific training to capture unique linguistic patterns in movie dialogues across different genres and decades. This enables effective detection of contextual abuse and subtle emotional expressions typical in cinematic dialogue, significantly improving practical accuracy for our longitudinal analysis framework.


\subsection{Framework}

    \begin{figure*}
    \centering
    \includegraphics[width=1\linewidth]{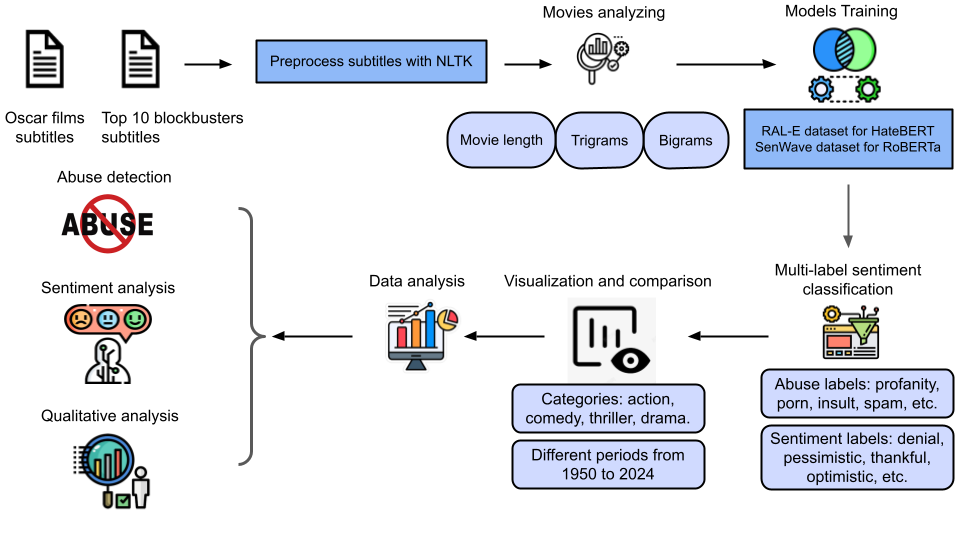}
    \caption{Our sentiment analysis and abusive detection framework for selected Hollywood movies from 1950 to 2024.}
    \label{fig:Framwork}
    \end{figure*}

We presented a framework for abuse detection and sentiment analysis of movie subtitles that includes several sequential steps, as shown in Figure \ref{fig:Framwork}. Our framework uses three datasets, including the movie subtitles, the SenWave dataset for sentiment analysis and RAL-E dataset for abuse detection for refining the RoBERTa models.

In Stage 1, we downloaded and curated movie subtitles. Oscar-nominated films and the top 10 Box Office films for each year were considered the most representative films, with categories shown in Table \ref{tab:filmcategories}. 

In Stage 2, we used the NLTK \citep{loper2002nltk} tool to remove the special phrases, such as hashtags, emotion symbols (emojis), stop words (e.g. “the”, “an”, “you”) for each movie subtitles during data processing. Stop words were also eliminated based on specific categories to ensure accurate bigrams and trigrams for subsequent analysis.

In Stage 3, we extracted the top ten bigrams and trigrams from movie subtitles for longitudinal analysis. Movie duration changes over the decades are also reported. Furthermore, the RAL-E and SenWave datasets were processed and cleaned for fine-tuning the BERT-based model at a later stage.

In Stage 4, we obtained two RoBERTa models from HuggingFace \citep{wolf2019huggingface} \footnote{\url{https://huggingface.co/docs/transformers/en/model_doc/roberta}}, fine-tuned with the SenWave and RAL-E datasets for sentiment analysis and abuse detection, respectively. The SenWave dataset enables comprehensive sentiment analysis by providing multi-label classification.

In Stage 5, we categorised the sentiments based on abuse analysis and sentiment analysis. For the abuse analysis, abusive emotions were categorised such as profanity, porn, insult, and spam. The sentiment analysis provides classifications such as denial, pessimistic, thankful, and optimistic to analyze positive and negative emotions more precisely.

In Stage 6, we analysed movie subtitles from different eras and categories using data visualisation, taking into account Oscar and Blockbuster categories along with the specific groups (action, comedy, drama, horror). Next, we compared the percentages of abusive text in each category and longitudinal sentiment polarity. Finally, we provided a subsequent comparison of a wider range of emotions, both positive and negative.

In Stage 7, we selected popular Oscar and Blockbuster movies, provided sentiment analysis and abuse detection on the entire movie timeline. This is reviewed by taking snapshots of dialogues for qualitative analysis.

 \subsection{Technical Details}
 
We utilise the state-of-the-art NLP techniques with pre-trained models based on RoBERTa \footnote{\url{https://huggingface.co/FacebookAI/roberta-base}}, specifically the HateBERT model \citep{caselli2020hatebert} optimised for detecting hate speech. We used the pre-trained HateBERT models from the Transformers library \footnote{\url{https://github.com/huggingface/transformers}}. Given the large dataset, we implemented batch processing to efficiently make predictions. Finally, we found that RoBERTa was more efficient in refining using the SenWave dataset.

 \section{Results}

 \subsection{Data analysis} 
 
\begin{figure*}
    \centering
    \begin{subfigure}[b]{\textwidth}
        \centering
        \includegraphics[height=1.5in]{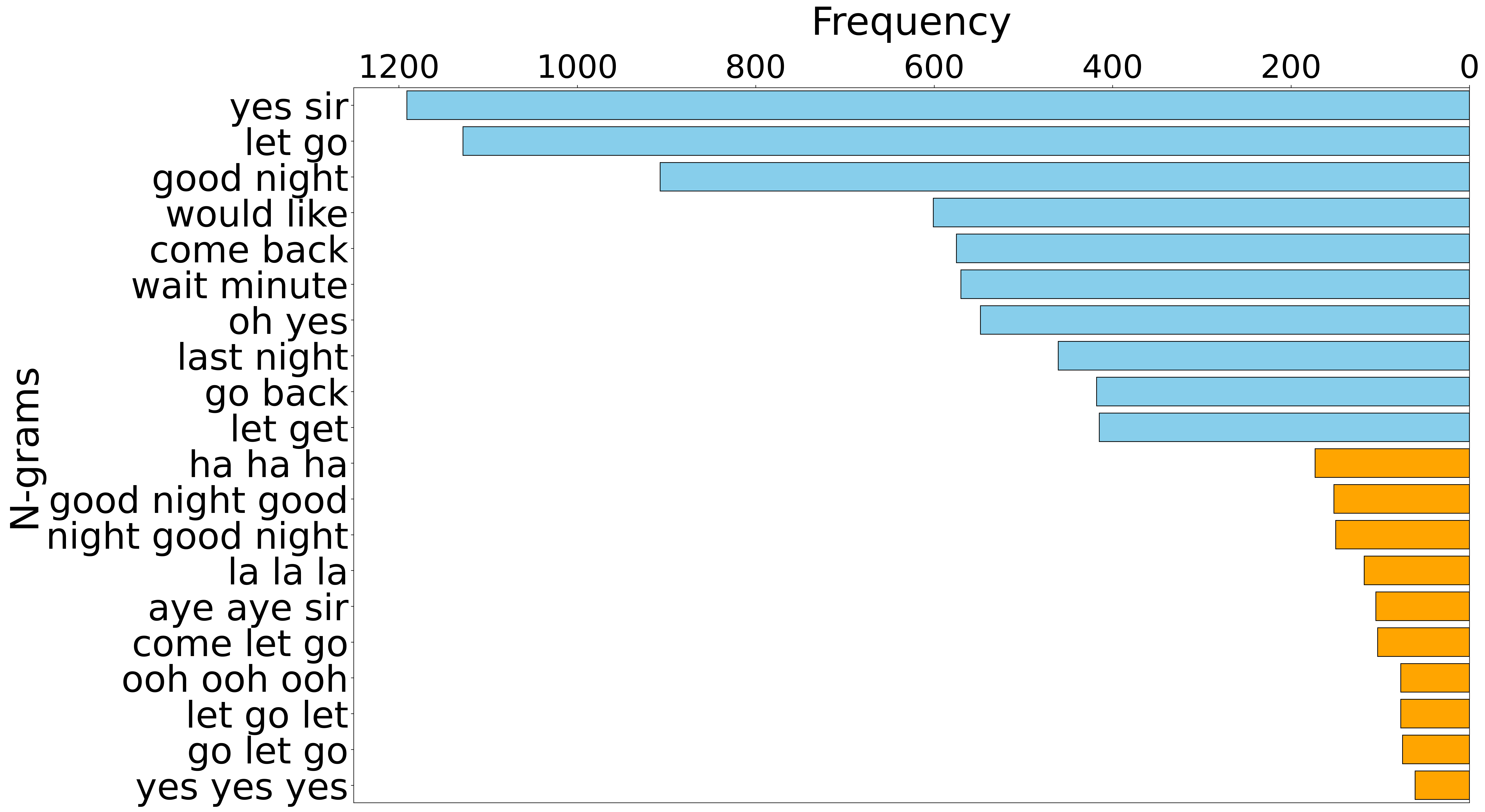}
        \caption{Ngram for 1950-1969}
    \end{subfigure}

    \begin{subfigure}[b]{\textwidth}
        \centering
        \includegraphics[height=1.5in]{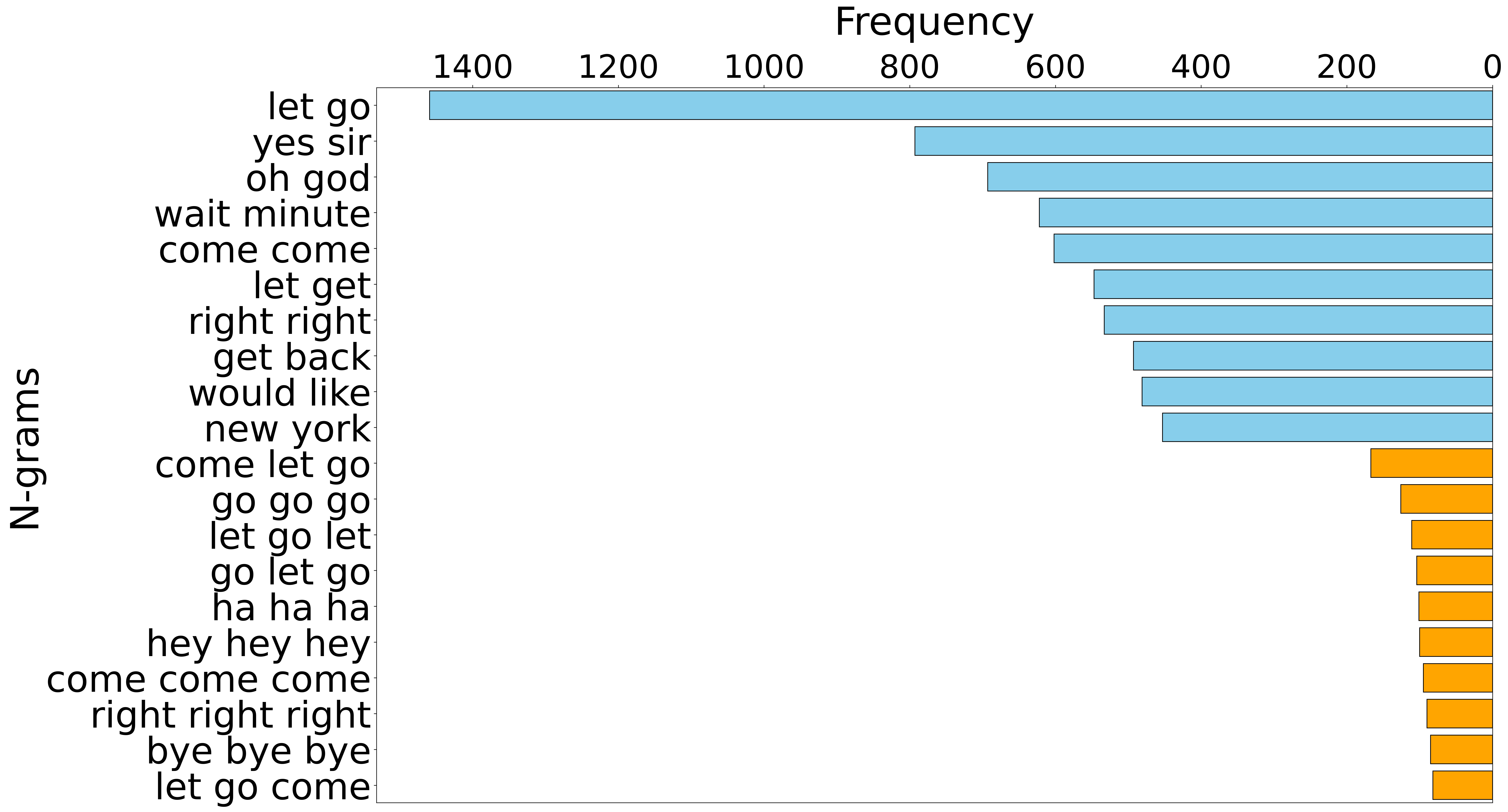}
        \caption{Ngram for 1970-1989}
    \end{subfigure}
    
    \begin{subfigure}[b]{\textwidth}
        \centering
        \includegraphics[height=1.5in]{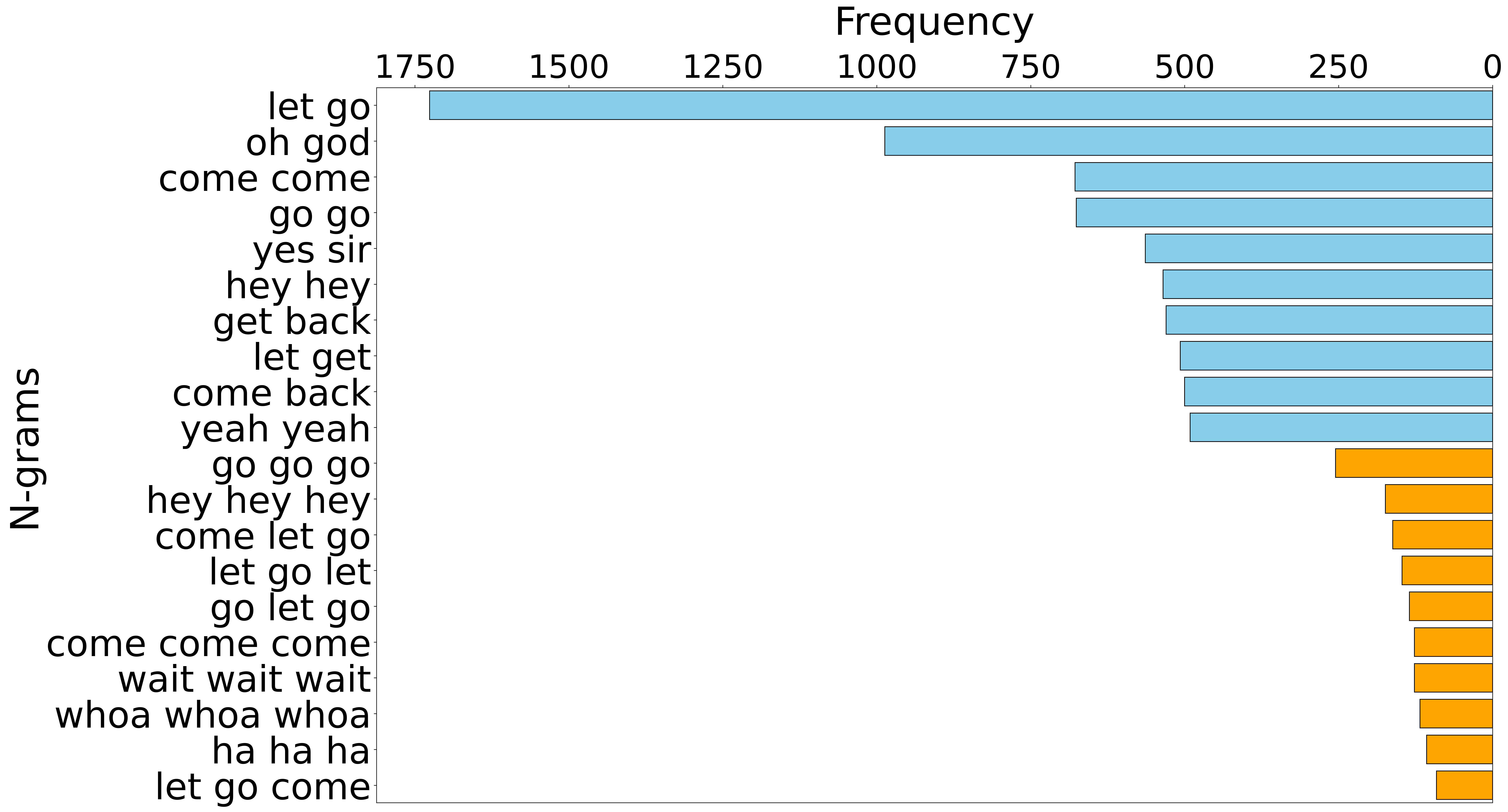}
        \caption{Ngram for 1990-2009}
    \end{subfigure}

    \begin{subfigure}[b]{\textwidth}
        \centering
        \includegraphics[height=1.5in]{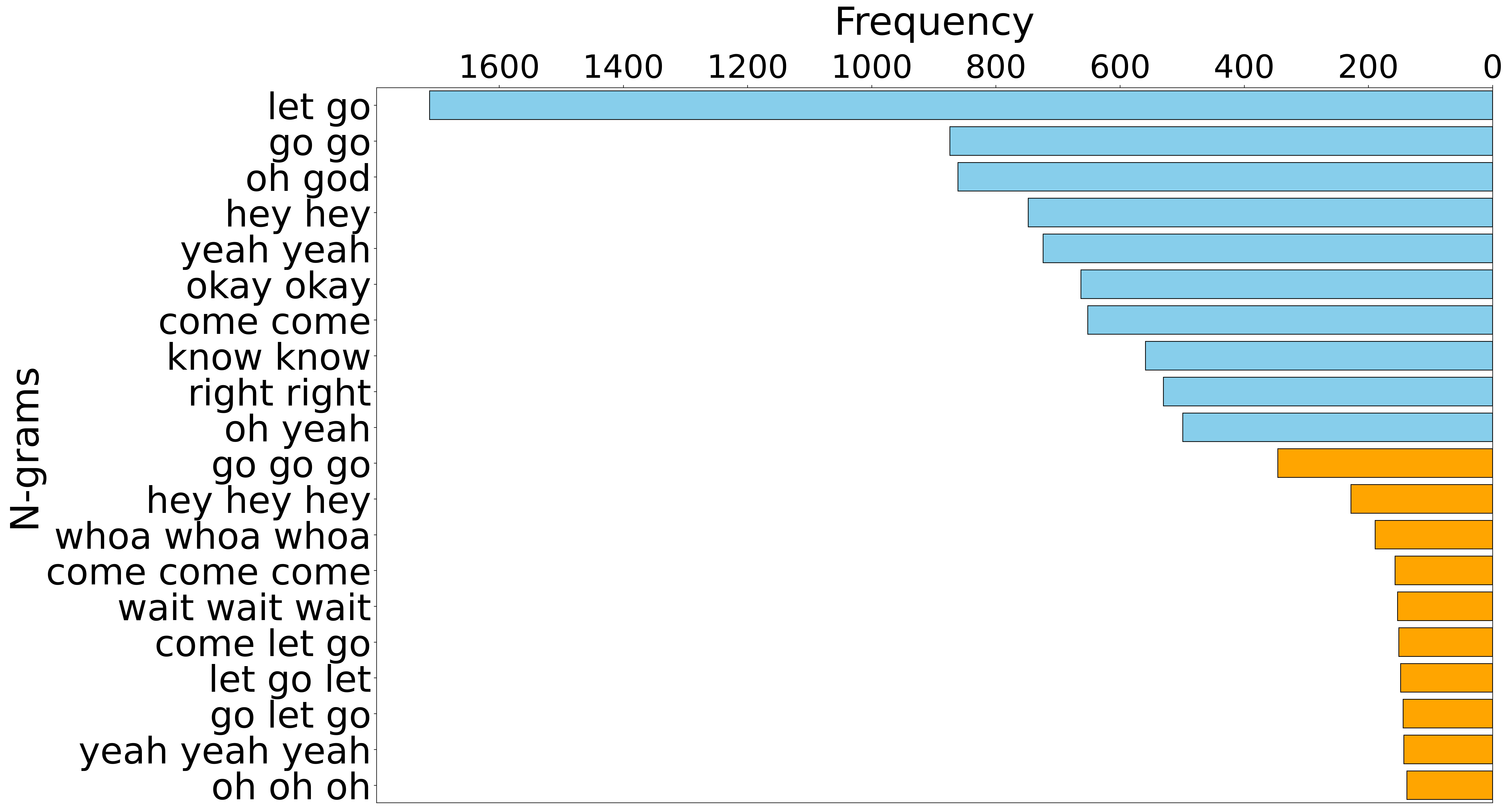}
        \caption{Ngram for 2010-2024}
    \end{subfigure}

    \caption{Top 10 bigrams and trigrams for different decades showing longitudinal changes.}
    \label{fig:ngramsdecadal}
\end{figure*}

We use n-gram analysis to outline the changes in the top twenty words of theme and sentiment in selected films over every 20 years. In Figure \ref{fig:ngramsdecadal}-Panel (a), we can observe that from 1950 to 1969, positive and colloquial phrases such as "yes sir", "let go" and "good night" were frequent, and these words usually appear in contexts that express affirmation or goodwill. In addition, "ha ha ha" and “good night good” indicate that the movies from this period are positive and affirmative emotions. Between 1970 and 1989 (Figure \ref{fig:ngramsdecadal})-Panel (b), we can see that common bigrams such as "let go" and "yes sir" still have a dominant position and "oh god" and "come come" have risen significantly. These changes reflect shock, surprise and action in the film. The appearance of "come let go" and "go go go"  shows an increase in the film's depiction of the action.  
In the period 1990-2009 (Figure \ref{fig:ngramsdecadal}-Panel c), we can see that "let go", "oh god" and "come come" still appear more frequently in bigrams than previously. The common trigrams such as "go go go" and "come let go" continue to appear, but "hey hey hey" rises in frequency. The rapid increase in word frequency indicates a significant increase in the length of subtitles in movies over the past two decades. The similarity of high-frequency words indicates that the main emotions of films during this period are the same as those of different periods in the past.
During the period 2010-2024 (Figure \ref{fig:ngramsdecadal}-Panel d), both bigrams and trigrams contain high-frequency particles, such as "hey hey", "yeah yeah", "whoa whoa whoa" and "oh oh oh". This indicates that emotions in films during this period were still predominantly positive, accompanied by more vocabulary that enhanced emotional expression.

\begin{figure}[h!]
    \centering
    \includegraphics[width=\linewidth]{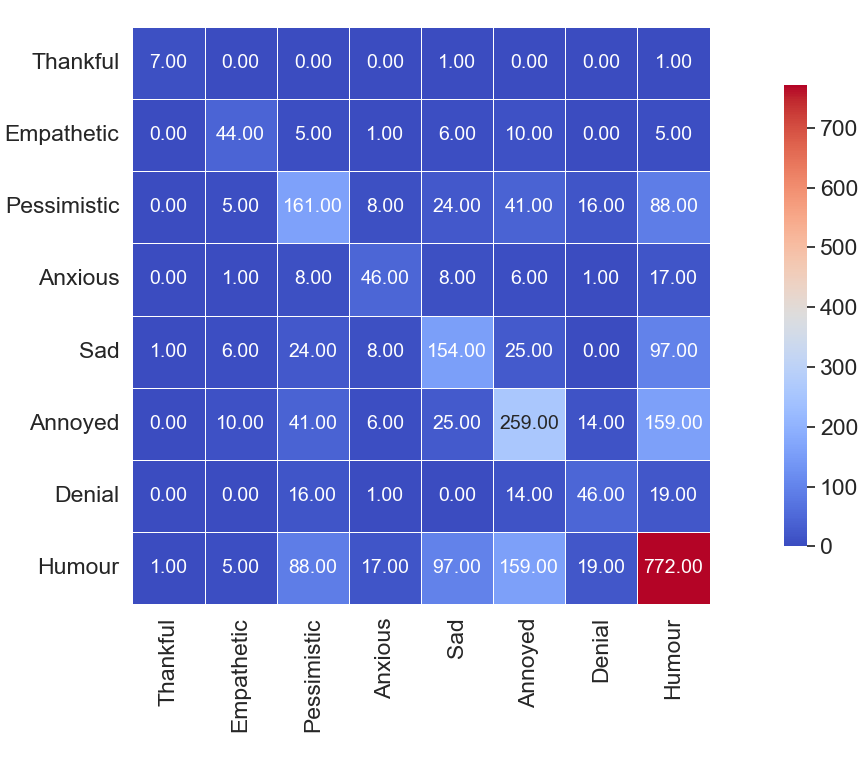}
    \caption{Heatmap about Sentiments Co-occurrence in movies during 75 years}
    \label{fig:sentiment co-occurrence}
\end{figure}

\subsection{Sentiment analysis}

\begin{table}[htbp]
\centering
\resizebox{\linewidth}{!}{
\begin{tabular}{lccccc}
\hline
\textbf{Model} & \textbf{Acc.} & \textbf{Prec.} & \textbf{Rec.} & \textbf{Micro F1} & \textbf{Ham. Loss} \\
\hline
TF-IDF + LR   & 0.44 & 0.62 & 0.57 & 0.59 & 0.118 \\
TF-IDF + SVM  & 0.45 & 0.64 & 0.56 & 0.60 & 0.115 \\
BERT-base     & 0.48 & 0.68 & 0.62 & 0.65 & 0.102 \\
RoBERTa-base  & \textbf{0.52} & \textbf{0.71} & \textbf{0.66} & \textbf{0.68} & \textbf{0.094} \\
\hline
\end{tabular}
}
\caption{Performance comparison of baseline models and RoBERTa model on SenWave dataset (English, multi-label).}
\label{senwavemetrics}
\end{table}

We evaluated the multi-label sentiment detection on the English subset of the SenWave dataset (10,000 tweets) and divided the dataset using iterative multilabel stratification into 70\% training, 10\% validation, and 20\% test, with a fixed random seed (42) for reproducibility. We provided the results using baseline \textcolor{black}{models based on TF-IDF vector representations \citep{das2023comparative} of the text, including unigrams and bigrams (ngram\_range=(1,2)), limited to the top 50,000 features by term frequency (max\_features=50,000).} Logistic Regression (TF-IDF + LR) and Linear Support Vector Machine (TF-IDF + LSVM) serve as classifiers.

As shown in Table~\ref{senwavemetrics}, the traditional baseline models perform modestly, with Subset Accuracy around 0.44–0.45 and Micro-F1 below 0.61. Fine-tuned BERT-base improves to 0.65 Micro-F1 with balanced Precision (0.68) and Recall (0.62). RoBERTa-base achieves the best overall results, reaching 0.68 Micro-F1, 0.52 Subset Accuracy, and the lowest Hamming Loss (0.094). Its higher recall (0.66) relative to BERT explains the gain, highlighting the advantage of domain-robust pre-trained models for multi-label sentiment classification.

\begin{figure}[h!]
    \centering
    \includegraphics[width=\linewidth]
    {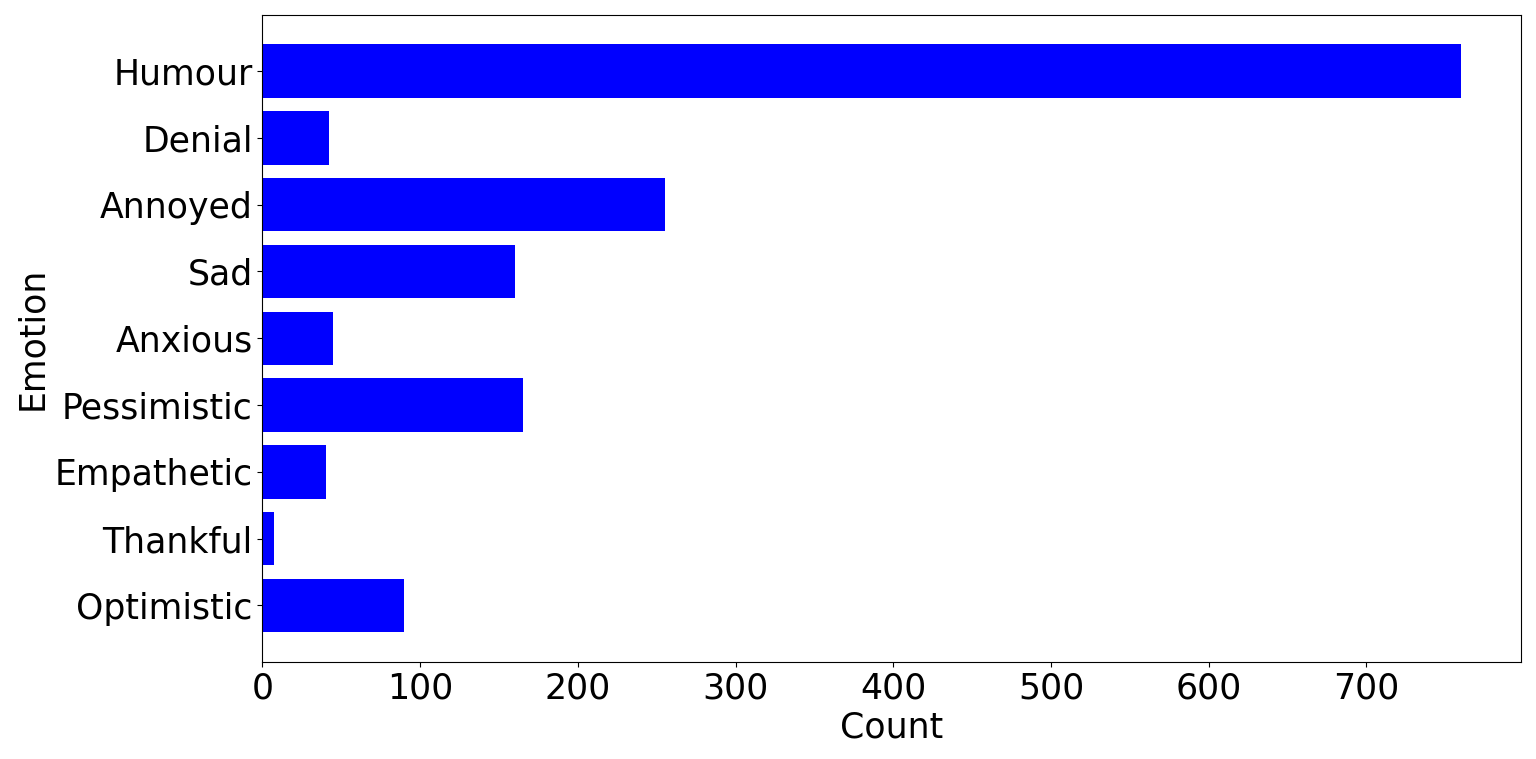}
    \caption{Counts of different emotions for Oscar-nominated and top ten blockbuster movies from 1950-2024.}
    \label{fig:counts_emotion}
\end{figure}

\begin{figure*}[h!]
    \centering
    \begin{subfigure}[b]{0.45\linewidth}
        \centering
        \includegraphics[width=\linewidth]{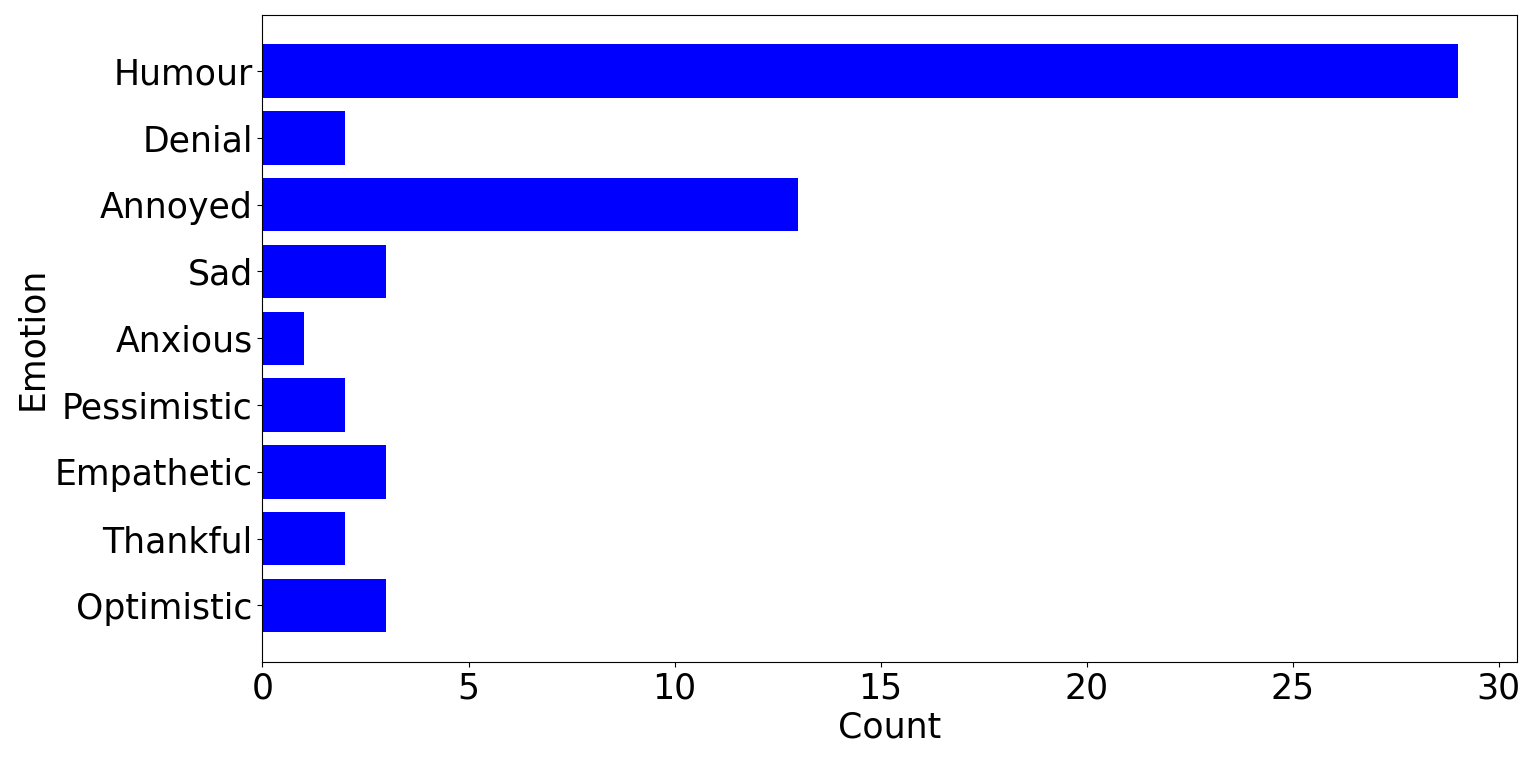}
        \caption{Action Category}
        \label{fig:action_category}
    \end{subfigure}
    \hfill
    \begin{subfigure}[b]{0.45\linewidth}
        \centering
        \includegraphics[width=\linewidth]{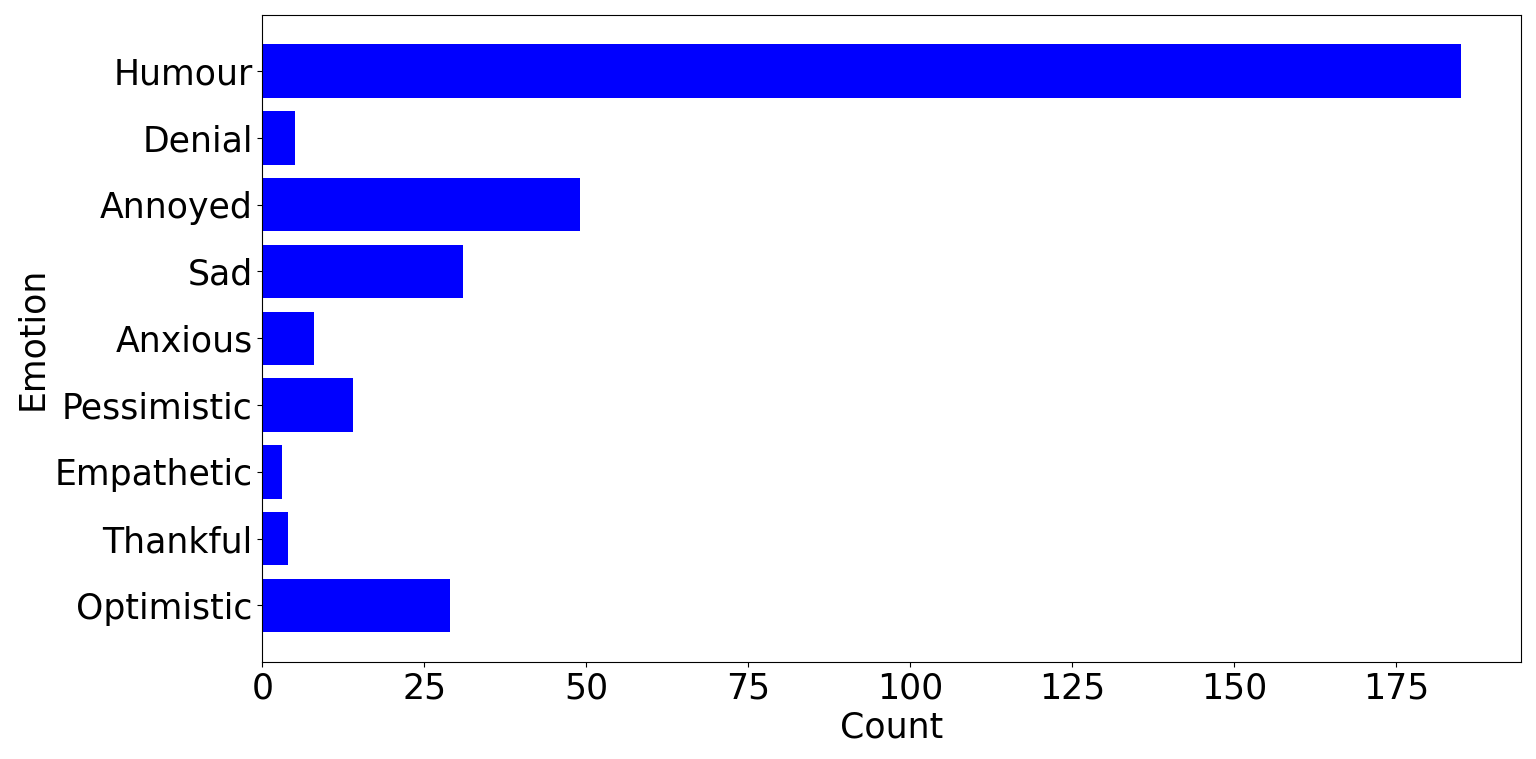}
        \caption{Comedy Category}
        \label{fig:comedy_category}
    \end{subfigure}
     
    \begin{subfigure}[b]{0.45\linewidth}
        \centering
        \includegraphics[width=\linewidth]{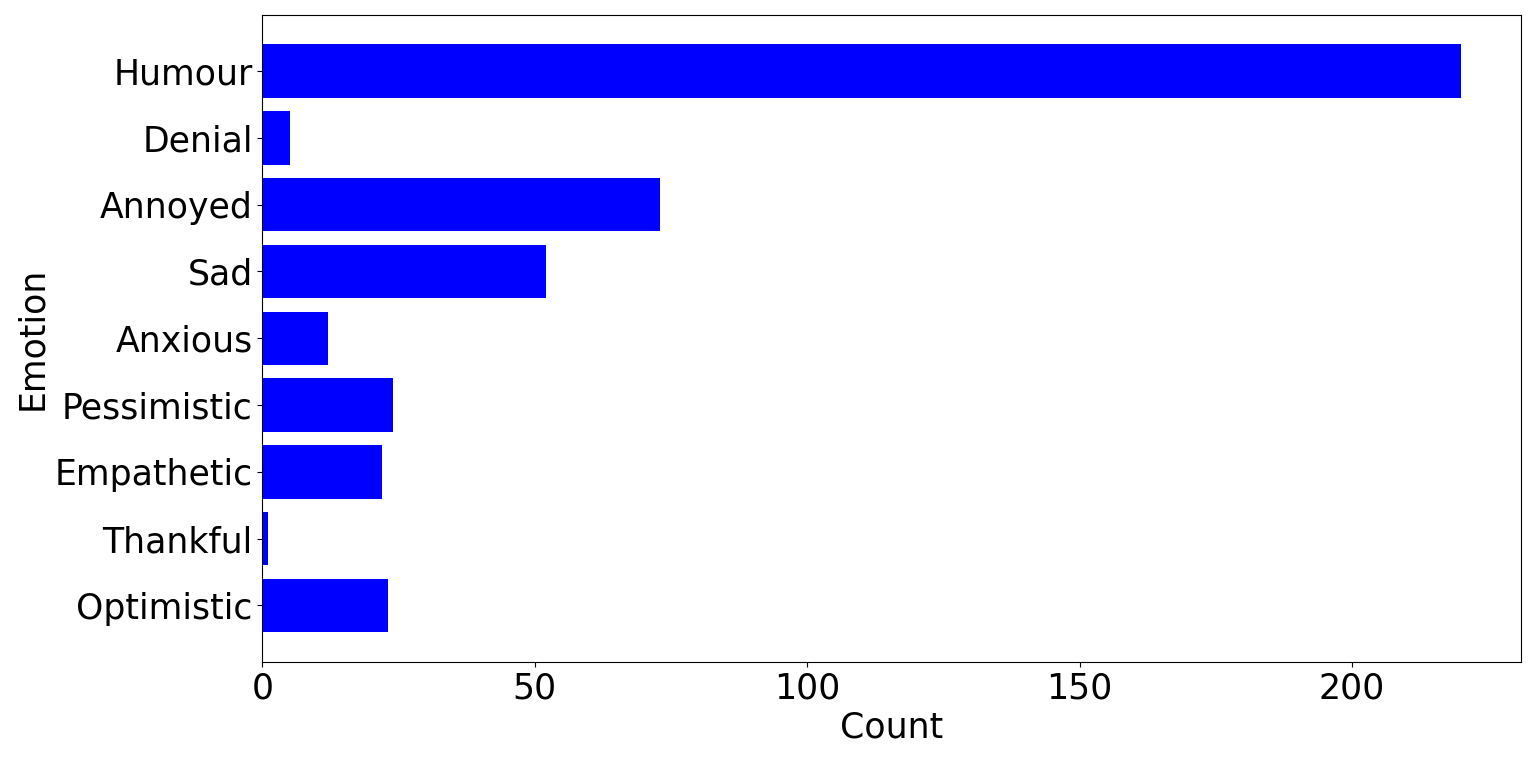}
        \caption{Drama Category}
        \label{fig:drama_category}
    \end{subfigure}
    \hfill
    \begin{subfigure}[b]{0.45\linewidth}
        \centering
        \includegraphics[width=\linewidth]{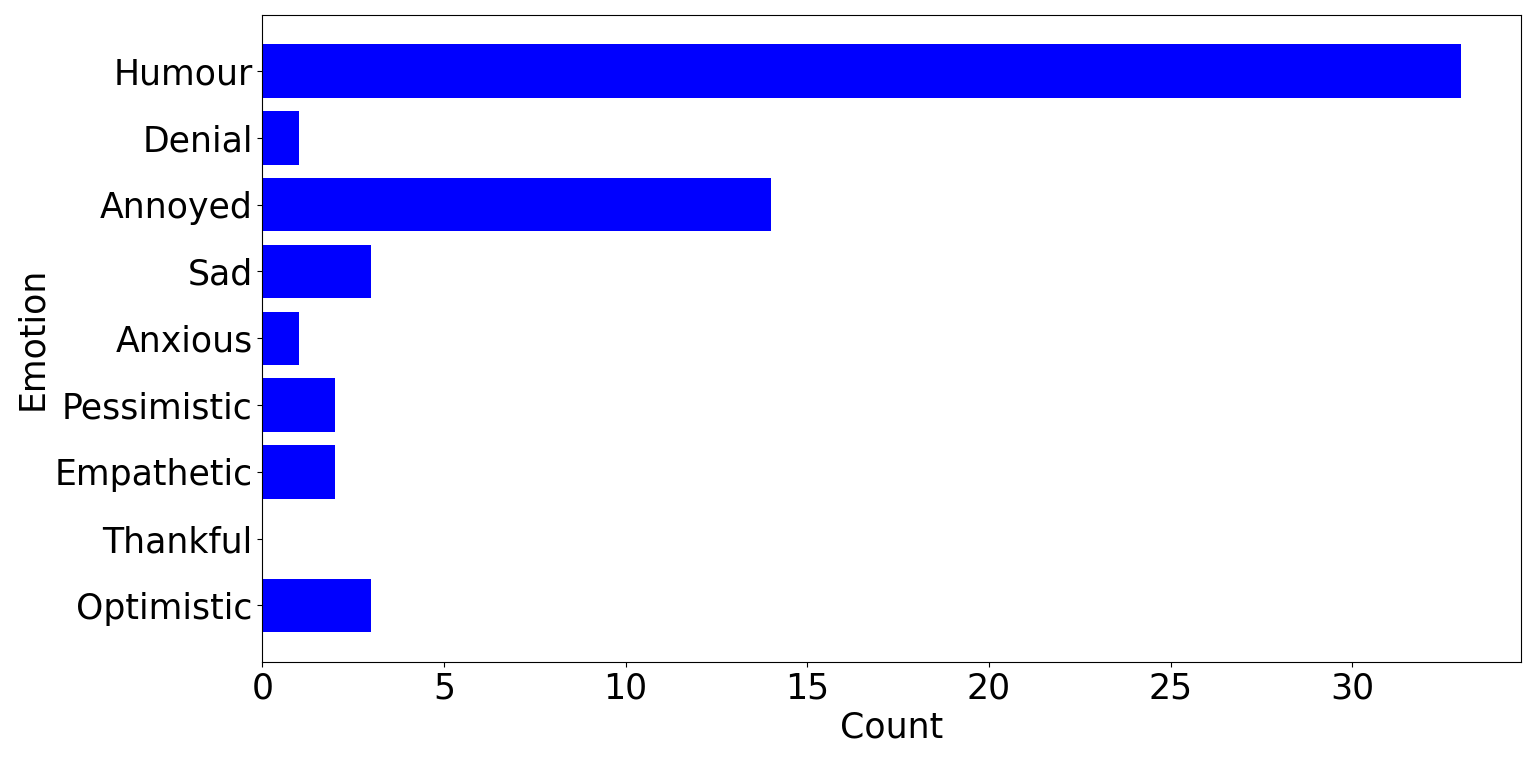}
        \caption{Thriller Category}
        \label{fig:thriller_category}
    \end{subfigure}
    
    \caption{Emotion Counts for Different Categories}
    \label{fig:emotion_counts_all}
\end{figure*}

Movies are generally highly expressive, and there is often no single emotion that can be expressed in a film. The different types of emotions captured from the subtitles depend on the type of film as well as the specific plot.   Figure \ref{fig:counts_emotion} illustrates the frequency of different emotions in movies detected by RoBERTa refined using SenWave data. We find that humour (joking) is the most frequent sentiment, which indicates that humour and lighthearted elements are essential in engaging audiences. This is followed by "annoyed", "sad",  "pessimistic", and optimistic sentiments, which include all the selected movies from 1950 to 2024. We note that these include Oscar-nominated movies and blockbusters in four major categories. Figure \ref{fig:emotion_counts_all} presents the results for the movies in the four major genres, including action, comedy, drama and thriller. We can observe that the humour sentiment is expressed the most in all categories, followed by "annoyed". The proposition of "sad" is different, where drama has the most and action and thriller have the least. There is a high number of "sad" sentiments in the "comedy" category.  Most movies would fall into two categories; for example, the movie "Annie Hall" is an example of a movie that would fall in both comedy and drama and "The Departed" would fall in both action and thriller categories. Therefore, the sentiments detected have overlaps across the categories. 

\begin{figure}[h!]
    \centering
    \includegraphics[width=\linewidth]{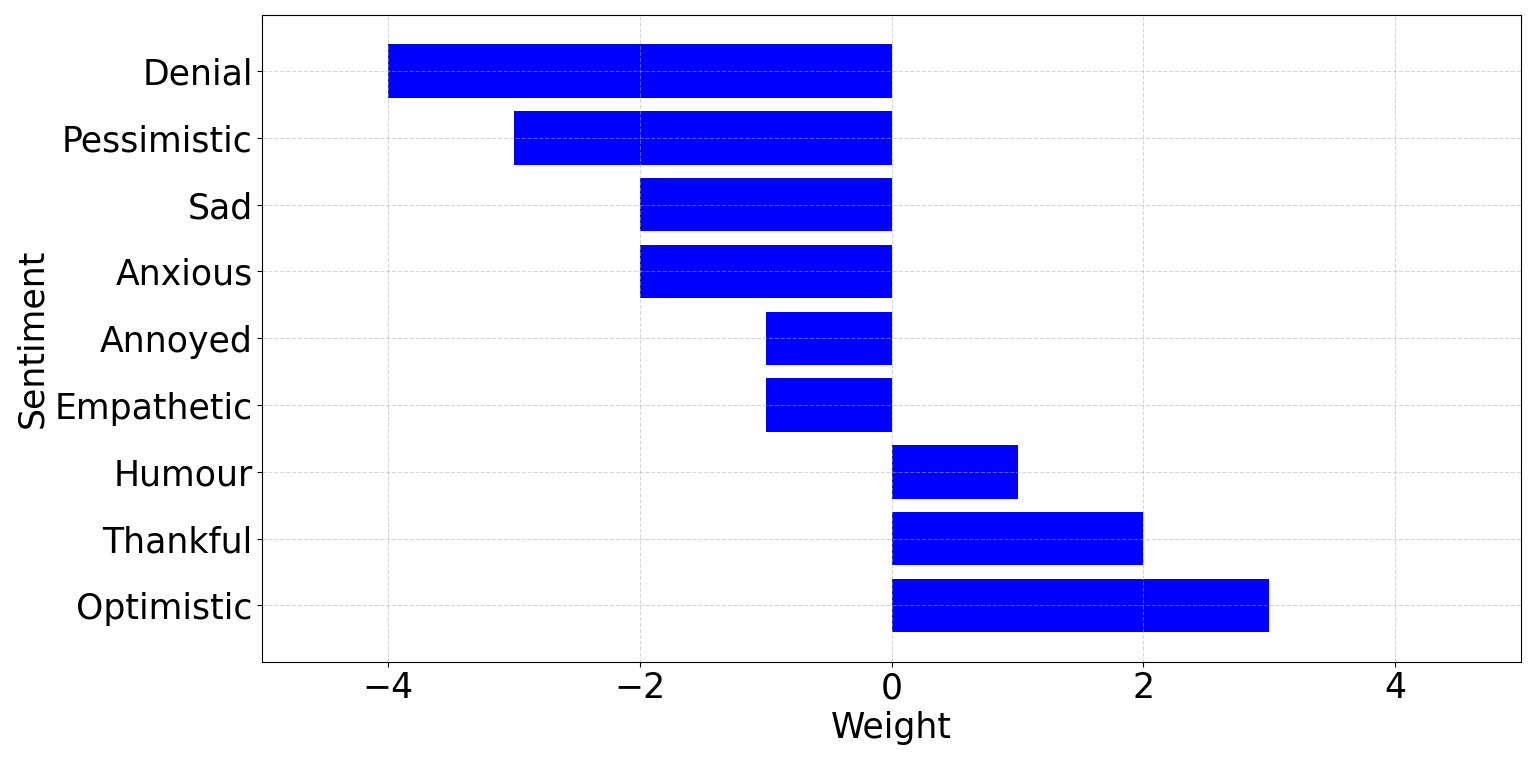}
    \caption{Sentiment Weights}
    \label{fig:sentiment_weight}
\end{figure}

We input the entire movie subtitle dataset into the RoBERTa model to generate word embeddings. A fully connected layer was then appended to the RoBERTa output, with the output dimensionality corresponding to the number of sentiment dimensions. \textcolor{black}{We acquired the sentiment weights ranging from -4 to 3 using an unconstrained linear activation function, enabling a direct interpretation of each sentiment dimension’s contribution to the overall polarity,} as depicted in Figure \ref{fig:sentiment_weight}. This demonstrates the contribution of different emotions to the overall affective polarity score, where positive numbers indicate positive emotions and negative numbers indicate negative emotions.

\begin{figure*}[h!]
    \centering
    \includegraphics[width=\linewidth]{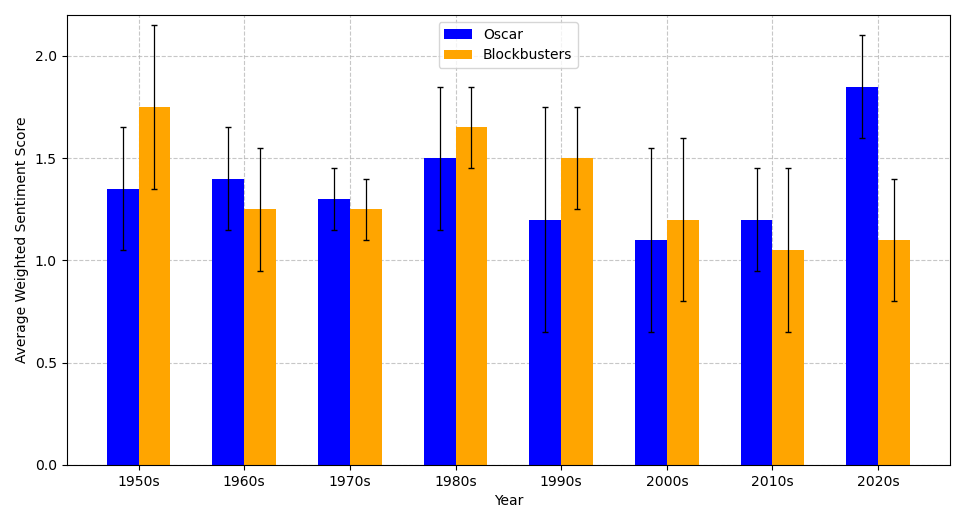}
    \caption{\textcolor{black}{Average weighted sentiment score trend for the given decades comparing Oscar and Blockbuster movies. Error bars represent the standard deviation (STD).}}
    \label{fig:ave_WSS}
\end{figure*}

Based on Figure \ref{fig:sentiment_weight}, we calculated the averag sentiment weight scores for every decade as shown in Figure \ref{fig:ave_WSS}. For blockbuster movies, the weighted sentiment scores are much higher during the 1950s, 1980s, and 1990s when compared to other periods, indicating that movies had richer emotional expressions during these three decades. In the case of Oscar-nominated movies, the average weighted sentiment score has been much higher than in other periods since 2020, indicating that the Oscars are gradually increasing their focus on films with richer emotional expressions. The standard deviation values in the bar chart are substantially higher during the 1950s, 1990s, and 2000s compared to other periods, indicating notable differences in emotional expression among films in these three periods.

Different themes and styles of movies cover different emotions, and multiple emotions can coexist in the same movie. Therefore, we obtain a heatmap of the co-occurrence of emotions in Figure \ref{fig:sentiment co-occurrence}, where the darker colours show higher co-occurrence frequency. We can observe that 'Humour' and 'Annoyed' have a high frequency that suggests many films use humour alongside anger, e.g. situations where jokes cause discomfort. Often, satire and dark humour are employed to show societal harsh realities. Similarly, 'Joking' and 'Sad' co-occurred 97 times, indicating that humour is often used to alleviate sadness, a characteristic commonly observed in tragicomic films. \textcolor{black}{Later decades showed more overlapping emotions such as humour mixed with annoyance, suggesting modern films use more complex emotional layering.}

At the same time, some low-frequency co-occurrence emotion phrases deserve our attention. "Thankful" and "Humour" \textcolor{black}{highlight a significant lack of immediate emotional transition or simultaneous appearance between these two states in the data.} The co-occurrence frequency of "Thankful" and "Pessimistic" is 0, which indicates that these two emotions hardly appear and in reality, thankfulness and pessimism are almost two opposite emotions and rarely appear in the same context.

\subsection{Abuse and Hate-Speech detection}

\begin{table*}[htbp]
\centering
\begin{tabular}{lcccc}
\hline
\textbf{Model} & \textbf{Accuracy} & \textbf{Precision} & \textbf{Recall} & \textbf{F1-score} \\
\hline
TF-IDF + Logistic Regression & 0.64 & 0.59 & 0.71 & 0.64 \\
TF-IDF + SVM                 & 0.66 & 0.70 & 0.61 & 0.65 \\
BERT-base       & 0.72 & \textbf{0.75} & 0.69 & 0.72 \\
HateBERT  & \textbf{0.77} & 0.74 & \textbf{0.81} & \textbf{0.77} \\
\hline
\end{tabular}
\caption{Performance comparison of baseline models and HateBERT model on RAL-E dataset.}
\label{ralemetrics}
\end{table*}

We compared three baseline models (TF-IDF + LR, TF-IDF + LSVM, and BERT-base) and the domain-adapted HateBERT model to evaluate the effectiveness of abuse detection. The RAL-E dataset, formulated as a binary classification task (abusive vs. non-abusive), is split into 70\% training, 10\% validation, and 20\% testing sets using stratified sampling to preserve label distribution.

\begin{figure}[h!]
    \centering
    \includegraphics[width=\linewidth]{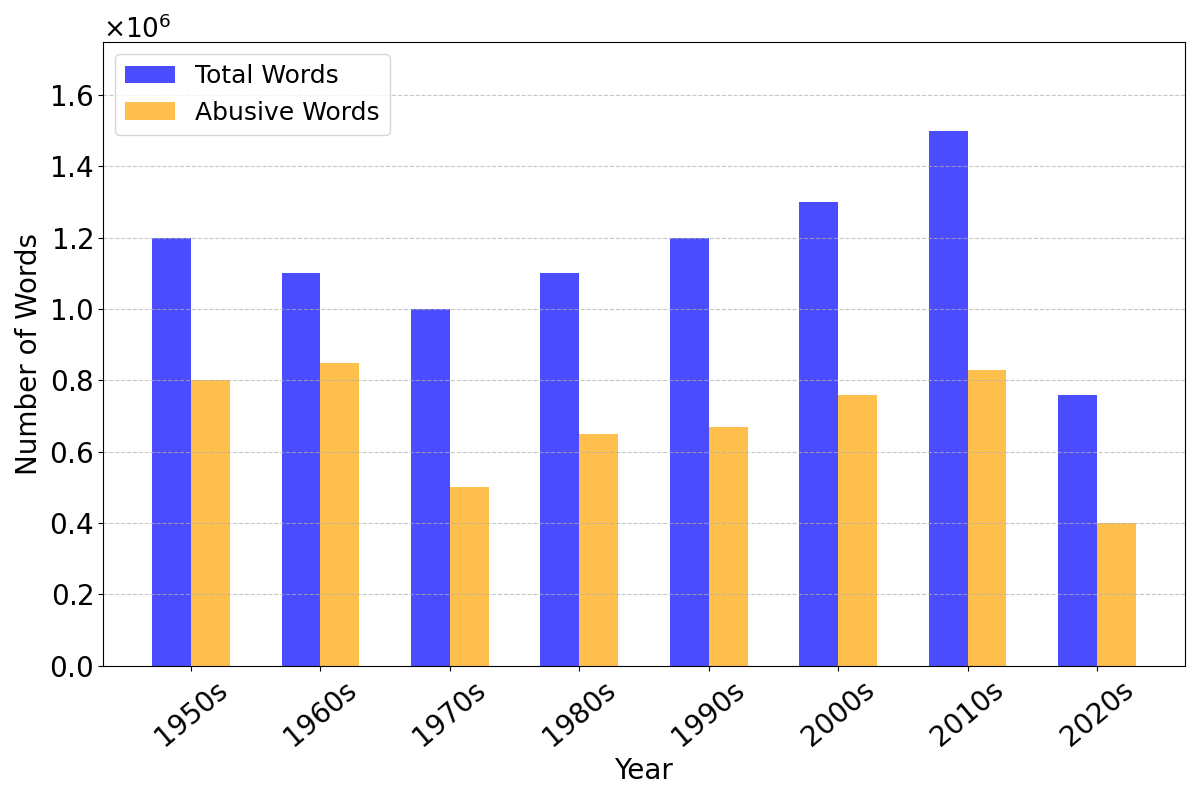}
    \caption{Total abusive words vs total words per decade.}    \label{fig:abusivewords}
\end{figure}
\begin{figure}[h!]
    \centering
    \includegraphics[width=\linewidth]{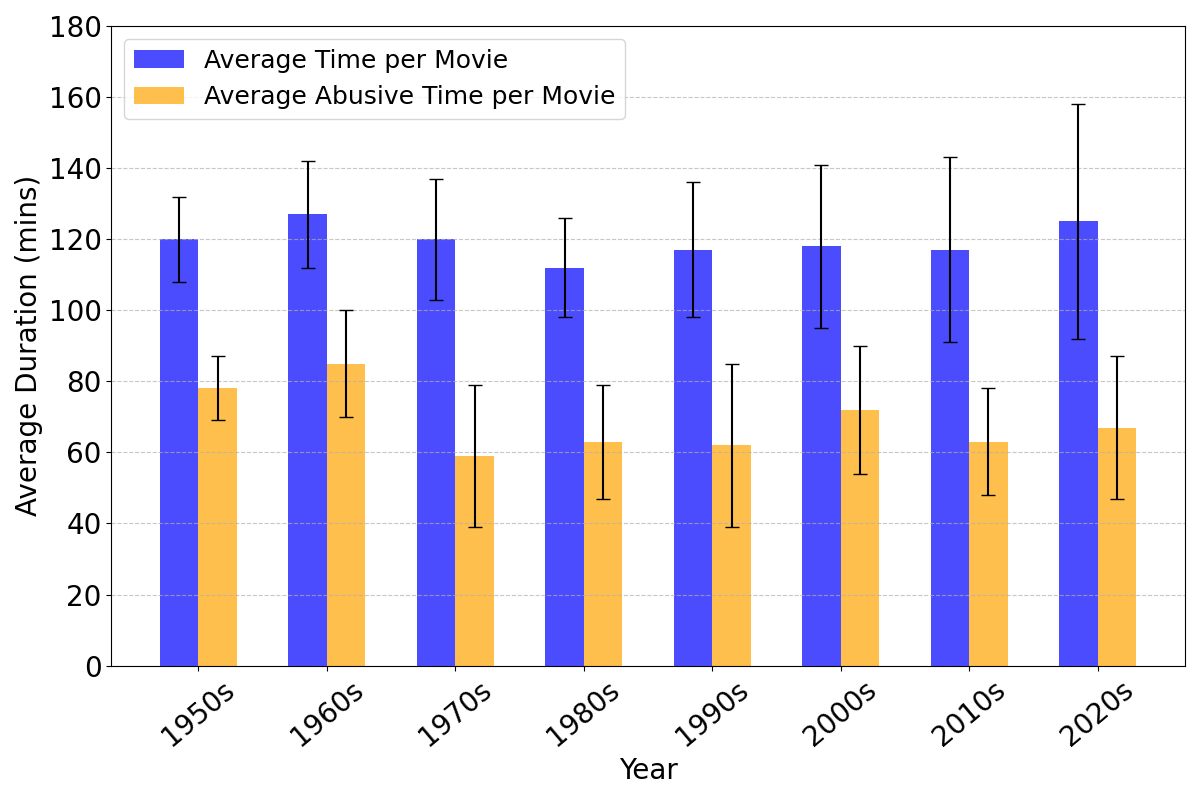}
    \caption{\textcolor{black}{Average movie time vs average abusive movie time per decade. Error bars represent the standard deviation (STD).}}
    \label{fig:abusivemovie}
\end{figure}

The results (Table~\ref{ralemetrics}) show that HateBERT achieved the highest accuracy score (0.77), and  TF-IDF + LR had the lowest (0.64). We focus on the F1-score as it accounts for the Precision and Recall. In terms of the F1-score, we find that HateBERT gave the best value (0.77) and TF-IDF + LR remained the weakest (0.64). Overall, these results highlight the limitations of traditional approaches and show that pre-trained language models, particularly those with domain-adaptive pretraining, provide notable gains in both overall accuracy and recall robustness. Figures \ref{fig:abusivewords} and \ref{fig:abusivemovie} present the results for abusive content, demonstrating how the movie subtitles have changed over the years, influenced by social-cultural, censorship, film genres, and external factors such as epidemics.

We utilise a look-up dictionary to count the number of abusive words. Figure \ref{fig:abusivewords} shows the trend of the number of words and abusive words in the film subtitles for each decade. We observe that the trend of the number of total words and the number of abusive words exhibits a greater degree of similarity across different decades. After the 1970s, the proportion of abusive words to total words in film subtitles across decades has not changed significantly. Such fluctuations may be attributed to changes in social tolerance for film content, the influence of censorship, and the emergence of new film genres. It is well known that abusive language was rare before 1980 but rose significantly afterwards, reflecting censorship and greater tolerance for controversial content.

Figure \ref{fig:abusivemovie} compares the average film durations in each decade with the durations of abusive content within them. The decline in abusive content post-1970s reflects shifting sociocultural norms and censorship policies. 

\begin{figure*}[h!]
    \centering
    \includegraphics[width=\linewidth]{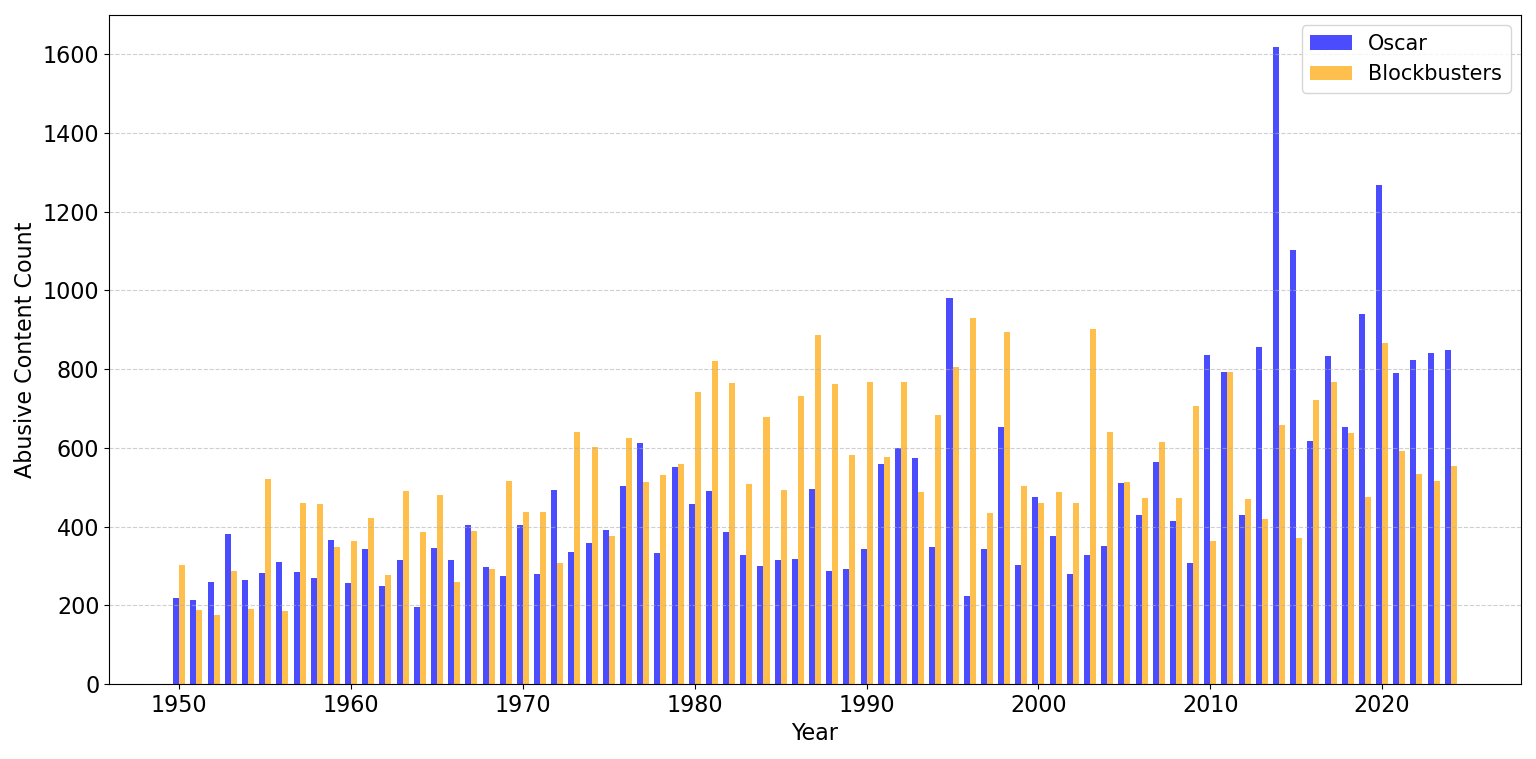}
    \caption{Trend of abusive content in Top 10 Blockbusters and  Oscar-Nominated movies (1950-2024).}
    \label{fig:Blockbuster}
\end{figure*}

Figure \ref{fig:Blockbuster} presents trends in the amount of violent content detected in Oscar-nominated films and top 10 blockbuster films between 1950 and 2024. Abusive content has shown a gradual increase since 1980, peaking in 2014, which may reflect the growth of the film industry and the changing social and cultural acceptance of abusive content. From 1950 to 1980, the amount of abuse in Oscar-nominated and top 10 blockbuster films was relatively small. This could be due to the more conservative social and cultural requirements of film content at the time. However,  abusive content gradually increased after the 1980s, and the amount of abuse in the top 10 blockbuster films almost surpassed that in Oscar-nominated movies before 2010, and has fallen significantly since then. This may indicate that Oscar-nominated films have become bolder and more diverse in subject and content, and the increase in abusive content may be an attempt to better represent complex social issues and human conflicts. Overall, we observe evolving trends in the representation and amount of abusive content in films over time, which reflect changes in social culture, the film industry, and audience preferences. Before 1980, abusive content was less and more restrained, and then increased with the diversification of film genres, especially in commercial films. Abusiveness in Oscar-nominated films has increased significantly since 2010, reflecting films' exploration of complex themes and commercialisation.

The years with the highest and lowest number of abuse for Oscars and Blockbuster movies were selected in Figure \ref{fig:Blockbuster}, and displayed the corresponding movie titles in Table \ref{table:films}.
Analysis of Oscar-nominated films shows a peak in abusive content in 2014 and 2020. Films like The Wolf of Wall Street and The Irishman use frequent expletives to heighten dramatic tension, emphasise the protagonists' volatile personalities, and create a gritty atmosphere. 12 Years a Slave and Dallas Buyers Club incorporate vulgarity to authentically portray the harsh realities and psychological distress of their characters. Joker, a psychological thriller, features radicalised language reflecting the protagonist's severe mental illness. Other films may use moderate expletives to enhance the atmosphere and aesthetic.
 
In contrast, the low use of abusive language in 1950 and 1964 Oscar-nominated films, such as All About Eve, Born Yesterday, Father of the Bride, King Solomon's Mines, Sunset Boulevard (1950), and Cleopatra, How the West Was Won, Lilies of the Field, Tom Jones, and America America (1964), can be attributed to their focus on drama, adventure, and family-oriented narratives. Strict censorship codes like the Hays Code restricted offensive content, and cultural norms emphasised traditional moral values, influencing filmmakers to avoid vulgar language to appeal to broader audiences.
 
Similarly, the low abusive language in 1951 and 1952 blockbusters, including African Queen, At War with the Army, David and Bathsheba, Show Boat, That's My Boy, Hans Christian Andersen, Jumping Jacks, Sailor Beware, and The Snows of Kilimanjaro, can be attributed to their focus on adventure, comedy, romance, and historical themes. These genres prioritised wholesome, family-friendly narratives, further aligning with the cultural and censorship standards of the time. 
 
In comparison, the increased use of abusive language in 1996 and 1998 blockbusters such as Independence Day, Mission Impossible, Ransom, The Rock, and Twister can be attributed to the action, thriller, and sci-fi genres, where high-stress situations often necessitate stronger language. Films like Lethal Weapon 4 and There’s Something About Mary cater to adult audiences with mature humour, allowing for greater flexibility in language as societal attitudes towards cinema language evolved.

\begin{figure*}[h!]
    \centering
    \includegraphics[width=\linewidth]{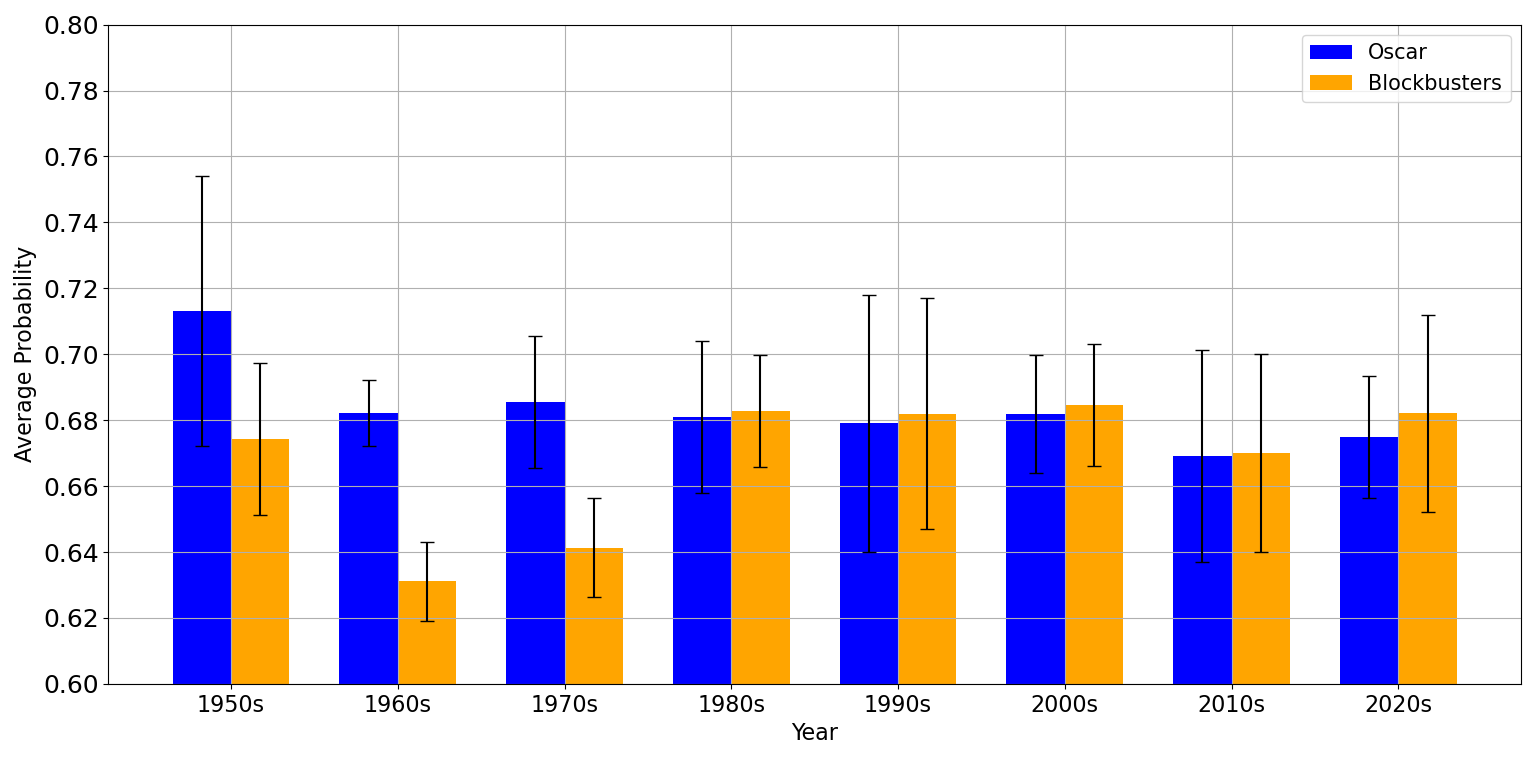}
    \caption{Average probability of abusive content by year.}
    \label{fig:AverageProbability}
\end{figure*}

 Figure \ref{fig:AverageProbability} shows how the probability of abusive content in Oscar-nominated films and the top 10 blockbuster films changed over the years between 1950 and 2024. The probability of abusive content is the amount of abusive content in a movie dialogue,  with 10-year timespan averaged. Although the average probability of abusive content fluctuates, Oscar-nominated films are far more likely to contain abusive content than the top 10 blockbuster films from the 1950s to 1970s. Afterwards, the probability of Oscar-nominated movies is almost equal to that of the top 10 blockbuster films.

\begin{figure}[h!]
    \centering
    \includegraphics[width=\linewidth]{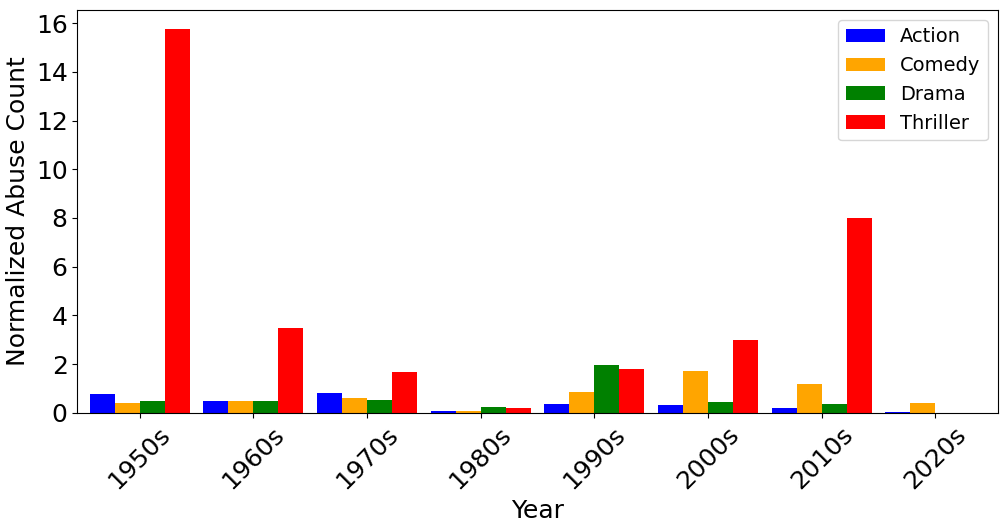}
    \caption{Normalised abuse count by category.}
    \label{fig:normalizedabusecount}
\end{figure}

 Figure \ref{fig:normalizedabusecount} presents the average abusive count (normalised count of abusive content per film) for each decade in four film categories (action, comedy, drama, thriller) between 1950 and 2024. The abusive count in action and drama films was stable and low across all decades, indicating relatively little abusive content in both categories. We note the unusually high abusive count of thrillers in the 1950s.

\begin{figure}[h!]
    \centering
    \includegraphics[width=\linewidth]{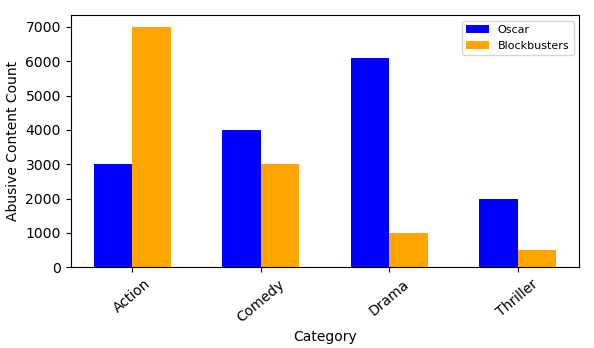}
    \caption{Comparison of frequency of abusive words for Oscar and Top-10 Blockbuster movies.}
    \label{fig:comparisonviolentwords}
\end{figure}

Figure \ref{fig:comparisonviolentwords} presents the frequency of abusive words between Oscar-winning movies and the top 10 blockbuster movies. Among action movies, the number of abusive words in the top 10 films is much higher than that of Oscar-winning films, nearly 7,000. This suggests that mainstream commercial action films often contain more abusive content, possibly to cater to the audience's need for excitement and strong emotions. The low number of abusive words in Oscar films, around 3,000, may indicate that Oscar films focus more on story depth and artistic expression rather than relying on violent elements.
Two categories of comedy films had relatively similar numbers of abusive words in this category, although Oscar films are slightly higher. This may indicate that there is less violence in comedies, both commercial and Oscar movies and that they are still primarily humorous.

\subsection{Case studies}

We presented some dialogues from four classic movies in Tables \ref{table:"The Departed" Dialogues}, \ref{table:"Annie Hall" Dialogues}, \ref{table:"The silence of the Lambs" Dialogues}, and \ref{table:"Chariots of Fire" Dialogues}, corresponding to the time points of the extreme values of sentiment polarity and abusive content count in Figures \ref{fig:The Departed}, \ref{fig:annie hall}, \ref{fig:the silence of the lambs}, and \ref{fig:chariots of fire}.

In Table \ref{table:"The Departed" Dialogues}, The Departed demonstrates a wide range of emotional tones. Time frame 00:27:32,684 to 00:27:34,777, \textit{"Yeah, that’s right, fancy. Are you a statie?"} expresses medium-level positive emotion, possibly indicating curiosity or mild amusement. However, in time frame 01:32:36,031 to 01:32:40,434, the tone shifts dramatically with \textit{"You pressure me to fear for my life and I will put a fucking bullet in your head...",} which conveys a high level of negative emotion, likely reflecting intense anger or threat. This intensity continues in the time frame 01:33:05,393 to 01:33:07,691 with \textit{"One of your guys is going to pop you,"} which also shows high negative emotion, suggesting violence or hostility. The sentiment remains negative from 01:33:10,365 to 01:33:12,663, where \textit{"As for running drugs, what the fuck are you doing?"} conveys frustration and strong disapproval. Time frame 01:33:12,834 to 01:33:16,429, \textit{"You don’t need the money or the pain in the ass. And they will catch you."} expresses medium-level negative emotion, indicating resignation or warning. Finally,  time frame 01:33:23,211 to 01:33:26,476, \textit{"Tell you the truth, I don’t need pussy anymore, either"} reveals a medium-level negative tone, possibly reflecting a sense of emotional detachment or bitterness. Overall, the emotional tone shifts sharply from medium positive to high negative, conveying strong hostility and frustration throughout the dialogue.

Annie Hall's sentiment shifts range from positive to negative in Table \ref{table:"Annie Hall" Dialogues}. In the beginning, from time frame  00:25:34,241 - 00:25:39,036, the phrase \textit{"You play well"} reflects a low level of positive emotion. Then, from 00:34:44,415 to 00:34:46,416, \textit{"Give me a kiss"} conveys a medium level of positive emotion. As the time progresses, the tone turns negative, with "Sometime you make me so mad!" from 00:44:05,142 to 00:44:09,270, and the following lines showing low-level negative emotion. This is followed by more negative sentiment in 00:52:36,820 to 00:52:40,197, criticising adult education with medium-level negativity. At 00:55:30,118 to 00:55:33,329, the statement about the Stones concert expresses low negative emotion. A shift occurs at 01:25:05,308 to 01:25:07,309, where \textit{"You know how wonderful you are"} introduces a high level of positive emotion. In the next few timestamps, the tone turns negative again, especially in 01:28:28,761 to 01:28:31,013 with references to \textit{"death,"} and in 01:29:13,097 to 01:29:15,432, with \textit{"difficult in life"} showing medium-level negativity. Overall, this dialogue showcases significant emotional fluctuations from positive to negative and back to positive, highlighting the complexity of interactions and the evolution of emotions.

In Table \ref{table:"The silence of the Lambs" Dialogues}, the emotional tone of The Silence of the Lambs fluctuates between positive and negative. In time frame  00:30:20,985 to 00:30:24,029, \textit{"Dr. Chilton does enjoy his petty torments"} expresses a low level of positive emotion, possibly hinting at amusement or satisfaction. This contrasts with the high level of negative emotion at  time frame 01:10:58,337 - 01:11:01,089, where the statement \textit{"They were slaughtering the spring lambs?"} carries a strong negative sentiment, likely reflecting shock or horror. At time frame 01:32:04,519 - 01:32:07,103, the mention of \textit{"Starling, we know who he is and where he is"} shows a low level of positive emotion, suggesting a sense of control or certainty. The emotional tone turns sharply negative again at time frame 01:35:33,477 - 01:35:35,311 with the phrase \textit{"I’d fuck me so hard,"} reflecting high negative emotion, likely in a disturbed or unsettling context. Finally, at time frame 01:45:32,034 - 01:45:35,077, \textit{"How the fuck should I know? Just get me out of here!"} expresses a high level of negative emotion, indicating frustration and desperation. Overall, the dialogue demonstrates a strong contrast between low positive and high negative emotions throughout these moments.

In Table \ref{table:"Chariots of Fire" Dialogues}, the time frame 00:10:44,140 - 00:10:48,020, "The enemy of one the enemy of all is" expresses a medium level of negative emotion, likely conveying a sense of conflict or tension. This is followed by another negative statement at time frame 00:10:59,260 - 00:11:01,700, where "They kicked me out of. Ring-a-ring o’Roses" also carries medium negative emotion, possibly suggesting rejection or frustration. The tone then shifts to positive, starting at time frame 00:45:48,800 - 00:45:51,200, with "Expecting great things, from all accounts" showing a low level of positive emotion, hinting at optimism or hope. At time frame 00:46:00,080 -00:46:02,600, "And good luck for this afternoon. Thank you." expresses medium-level positive emotion, offering encouragement and good wishes. Finally, at time frame 01:25:59,280 - 01:26:03,760, "Then let’s hope that’s wise enough to give you room for manoeuvre" expresses a low level of positive emotion, suggesting cautious optimism or hope for flexibility. Overall, the emotional tone moves from medium negative to positive, reflecting shifts in expectations and interactions.

\begin{table*}[h!]
    \centering
    \begin{tabular}{|c|c|p{8cm}|}
        \hline
        \textbf{Year} & \textbf{Category} & \textbf{Film Names} \\
        \hline
        1950 & Oscar & All About Eve, Born Yesterday, Father of the Bridge, King Solomon's Mines, Sunset Boulevard \\
        \hline
        1964 & Oscar & Cleopatra, How the West Was Won, Lilies of the Field, Tom Jones, America America \\
        \hline
        2014 & Oscar & 12 Years a Slave, American Hustle, Captain Phillips, Dallas Buyers Club, Gravity, Her, Nebraska, Philomena, The Wolf of Wall Street \\
        \hline
        2020 & Oscar & Parasite, Ford v Ferrari, The Irishman, Jojo Rabbit, Joker, Little Women, Marriage Story, 1917, Once Upon a Time in Hollywood \\
        \hline
        1951 & Blockbusters & African Queen, At War with the Army, David and Bathsheba, Show Boat, Thats My Boy, The Great Caruso \\
        \hline
        1952 & Blockbusters & Hans Christian Andersen, Jumping Jacks, Sailor Beware, The Snows of Kilimanjaro \\
        \hline
        1996 & Blockbusters & 101 Dalmatians, Independence Day, Mission Impossible, Ransom, Space Jam, The Hunchback of Notre Dame, The Nutty Professor, The Rock, Twister \\
        \hline
        1998 & Blockbusters & A Bug's Life, Armageddon, Deep Impact, Dr.Dolittle, Godzilla, Lethal Weapon 4, Mulan, There's Something About Mary \\
        \hline
    \end{tabular}
    \caption{Oscar and Blockbusters Film Lists by Year and Category}
    \label{table:films}
\end{table*}

\begin{table*}[h!]
    \centering
    \scalebox{0.85}{
    \begin{tabular}{|c|c|p{4cm}|c|c|}
        \hline
        \textbf{Start Time} & \textbf{End Time} & \textbf{Text} & \textbf{Sentiment} & \textbf{Abusive Level} \\
        \hline
        00:27:32,684 & 00:27:34,777 & Yeah, that's right, fancy.
Are you a statie? & Positive & Medium \\ 
        \hline
        01:32:36,031 & 01:32:40,434 & You pressure me to fear for my life and
I will put a fucking bullet in your head... & Negative & High \\ 
        \hline        
        01:33:05,393 & 01:33:07,691 & One of your guys is gonna pop you. & Negative & High \\ 
        \hline
        01:33:10,365 & 01:33:12,663 & As for running drugs,
what the fuck are you doing? & Negative & High \\ 
        \hline
        01:33:12,834 & 01:33:16,429 & You don't need the money or the pain
in the ass. And they will catch you. & Negative & Medium \\ 
        \hline
        01:33:23,211 & 01:33:26,476 & Tell you the truth,
I don't need pussy anymore, either. & Negative & Medium \\
    \hline
    \end{tabular}}
    \caption{Sentiment Analysis and Abusive Level Detection of "The Departed" Dialogues}
    \label{table:"The Departed" Dialogues}
\end{table*}

\begin{table*}[h!]
    \centering
    \scalebox{0.85}{
    \begin{tabular}{|c|c|p{4cm}|c|c|}
        \hline
        \textbf{Start Time} & \textbf{End Time} & \textbf{Text} & \textbf{Sentiment} & \textbf{Abusive Level} \\
        \hline
        00:25:34,241 & 00:25:39,036 & You say, "You play well" and then
right away I have to say, "You play well." & Positive & Low \\ 
        \hline
        00:34:44,415 & 00:34:46,416 & Hey, listen. Give me a kiss. & Positive & Medium \\
        \hline
        00:44:05,142 & 00:44:09,270 & "Oh, Marie!
Sometime you make me so mad!"
& Negative & Low \\ 
        \hline
        00:44:09,438 & 00:44:14,651 & They scream at that! Write me something
like that. A French number. Can you do it? & Negative & Low \\ 
        \hline
        00:52:36,820 & 00:52:40,197 & Adult education is such junk.
The professors are so phoney. & Negative & Medium \\ 
        \hline
        00:55:30,118 & 00:55:33,329 & I was at the Stones concert
when they killed that guy. & Negative & Low \\ 
        \hline
        01:25:05,308 & 01:25:07,309 & You know how wonderful you are. & Positive & High \\ 
        \hline
        01:28:28,761 & 01:28:31,013 & You only give me books
with "death" in the title. & Negative & Medium \\ 
        \hline
        01:29:13,097 & 01:29:15,432 & ...because it's real difficult in life. & Negative & Medium \\
        \hline
    \end{tabular}}
    \caption{Sentiment Analysis and Abusive Level Detection of "Annie Hall" Dialogues}
    \label{table:"Annie Hall" Dialogues}
\end{table*}

\begin{table*}[h!]
    \centering
    \scalebox{0.85}{
    \begin{tabular}{|c|c|p{4cm}|c|c|}
        \hline
        \textbf{Start Time} & \textbf{End Time} & \textbf{Text} & \textbf{Sentiment} & \textbf{Abusive Level} \\ 
        \hline
        00:30:20,985 & 00:30:24,029 & Dr. Chilton does enjoy his petty torments. & Positive & Low \\ 
        \hline
        01:10:58,337 & 01:11:01,089 & They were slaughtering the spring lambs? & Negative & High \\ 
        \hline
        01:32:04,519 & 01:32:07,103 & Starling. Starling,
we know who he is and where he is. & Positive & Low \\ 
        \hline
        01:35:33,477 & 01:35:35,311 & I'd fuck me so hard. & Negative & High \\ 
        \hline
        01:45:32,034 & 01:45:35,077 & How the fuck should I know?
Just get me out of here! & Negative & High \\ 
        \hline
    \end{tabular}}
    \caption{Sentiment and Abusive Level Analysis of "The Silence of the Lambs" Dialogues}
    \label{table:"The silence of the Lambs" Dialogues}
\end{table*}

\begin{table*}[h!]
    \centering
    \scalebox{0.85}{
    \begin{tabular}{|c|c|p{4cm}|c|c|}
        \hline
        \textbf{Start Time} & \textbf{End Time} & \textbf{Text} & \textbf{Sentiment} & \textbf{Abusive Level} \\ 
        \hline
        00:10:44,140 & 00:10:48,020 & ♪ The enemy of one the enemy of all is & Negative & Medium \\ 
        \hline
        00:10:59,260 & 00:11:01,700 & They kicked me out of. Ring-a-ring o'Roses. & Negative & Medium \\
        \hline
        00:45:48,800 & 00:45:51,200 & Expecting great things, from all accounts. & Positive & Low \\
        \hline
        00:46:00,080 & 00:46:02,600 & And good luck for this afternoon. Thank you. & Positive & Medium \\
        \hline
        01:25:59,280 & 01:26:03,760 & Then let's hope that's wise enough
to give you room for manoeuvre. & Positive & Low \\ 
        \hline
    \end{tabular}}
    \caption{Sentiment Analysis and Abusive Level Detection of "Chariots of Fire" Dialogues}
    \label{table:"Chariots of Fire" Dialogues}
\end{table*}

\subsubsection{Take some examples to conduct sentiment analysis and abusive detection for the entire movie}

\begin{figure*}[h!]
    \centering
    \includegraphics[width=\linewidth]{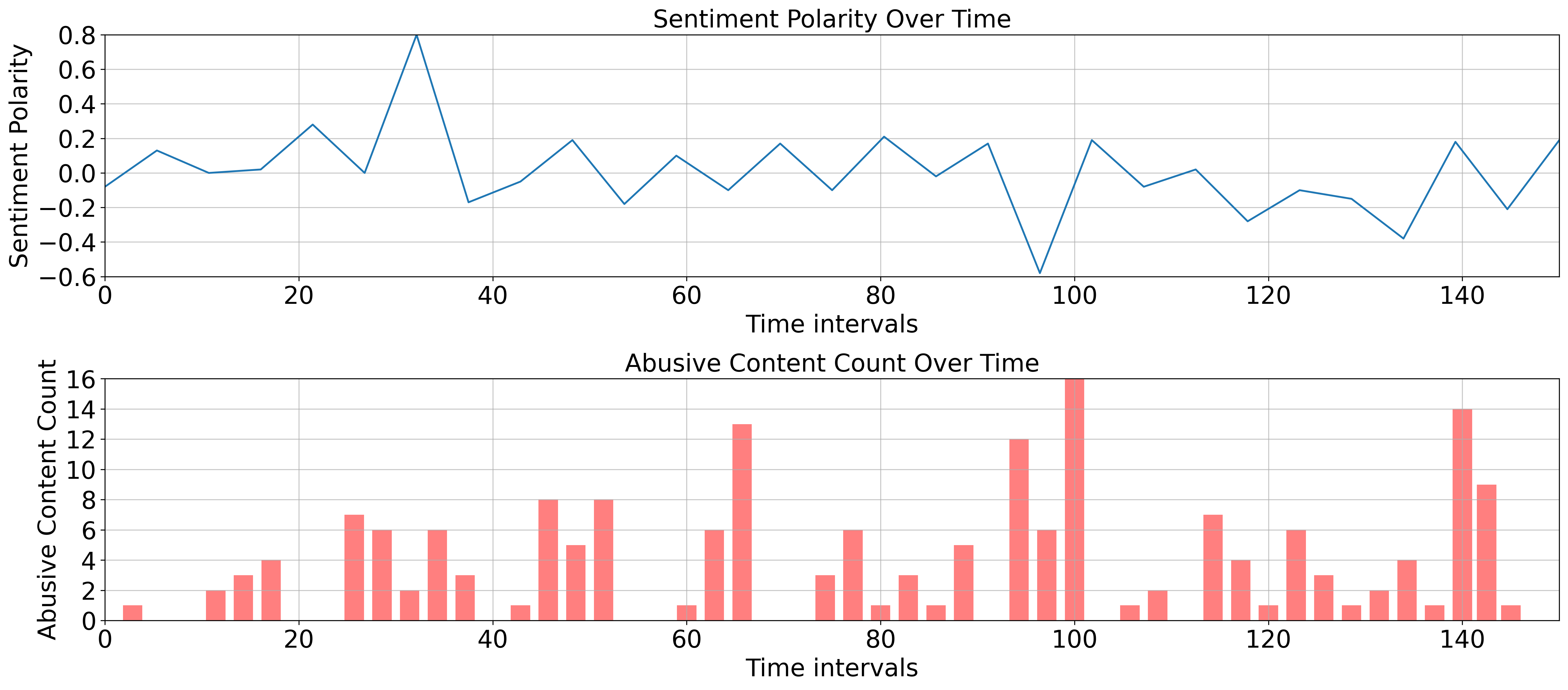}
    \caption{Take "The Departed" as an example to conduct sentiment analysis and abusive detection for the entire movie}
    \label{fig:The Departed}
\end{figure*}

\begin{figure*}[h!]
    \centering
    \includegraphics[width=\linewidth]{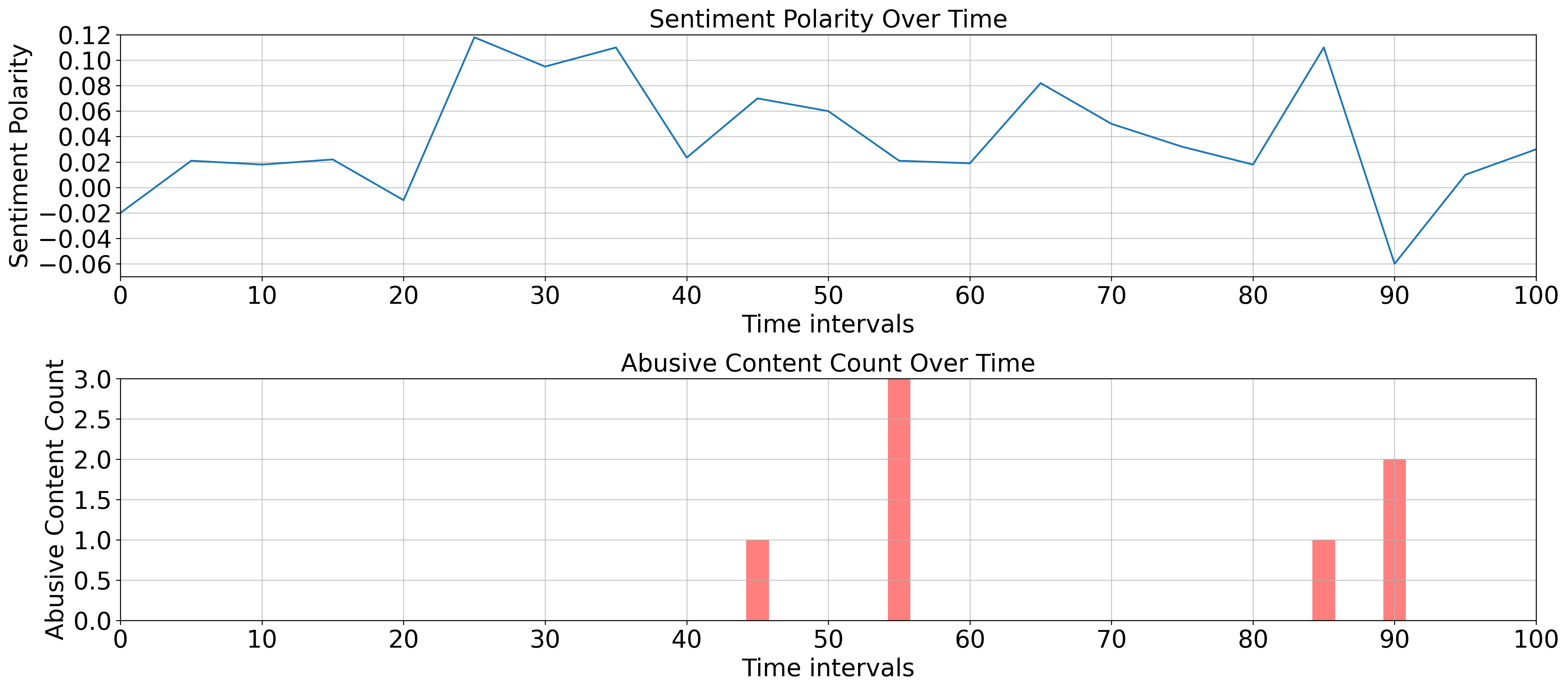}
    \caption{Take "Annie Hall" as an example to conduct sentiment analysis and abusive detection for the entire movie}
    \label{fig:annie hall}
\end{figure*}

\begin{figure*}[h!]
    \centering
    \includegraphics[width=\linewidth]{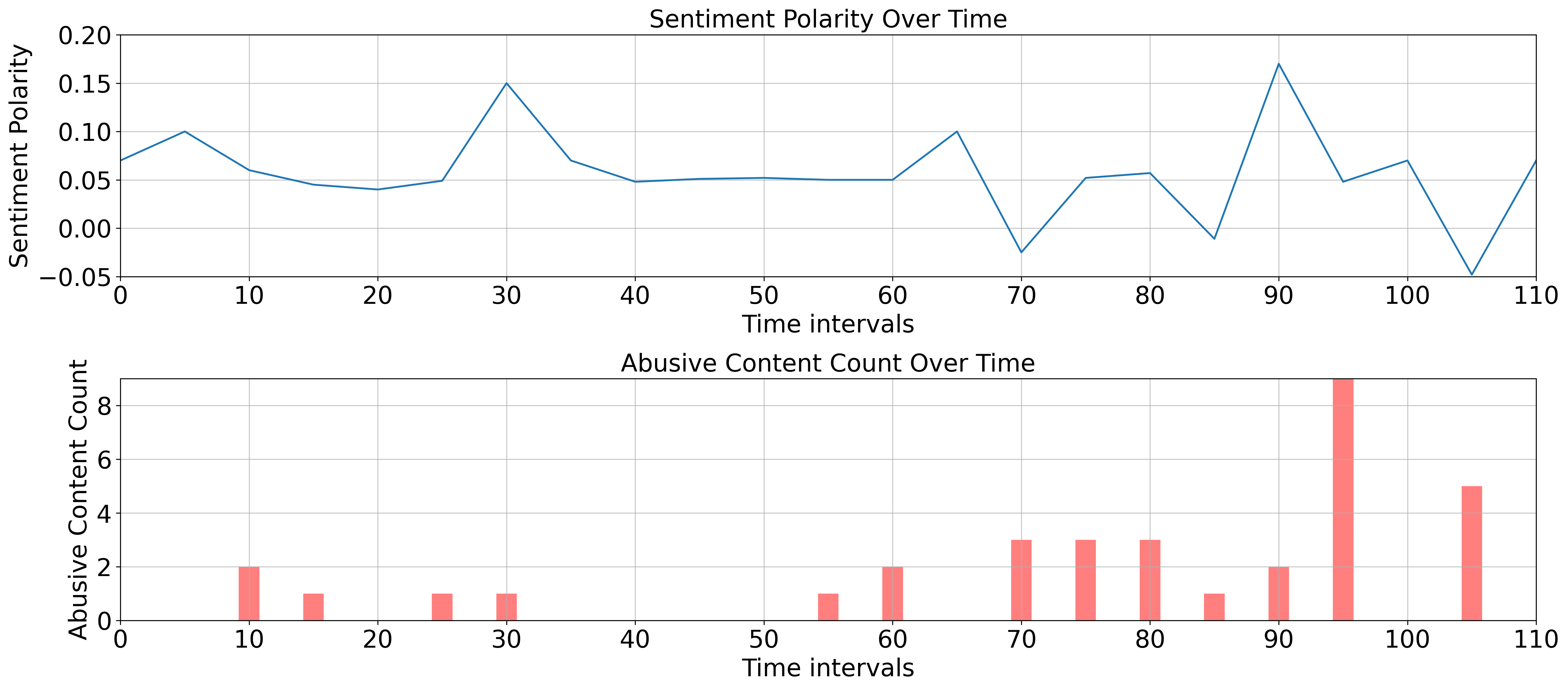}
    \caption{Take "the silence of the lambs" as an example to conduct sentiment analysis and abusive detection for the entire movie}
    \label{fig:the silence of the lambs}
\end{figure*}

\begin{figure*}[h!]
    \centering
    \includegraphics[width=\linewidth]{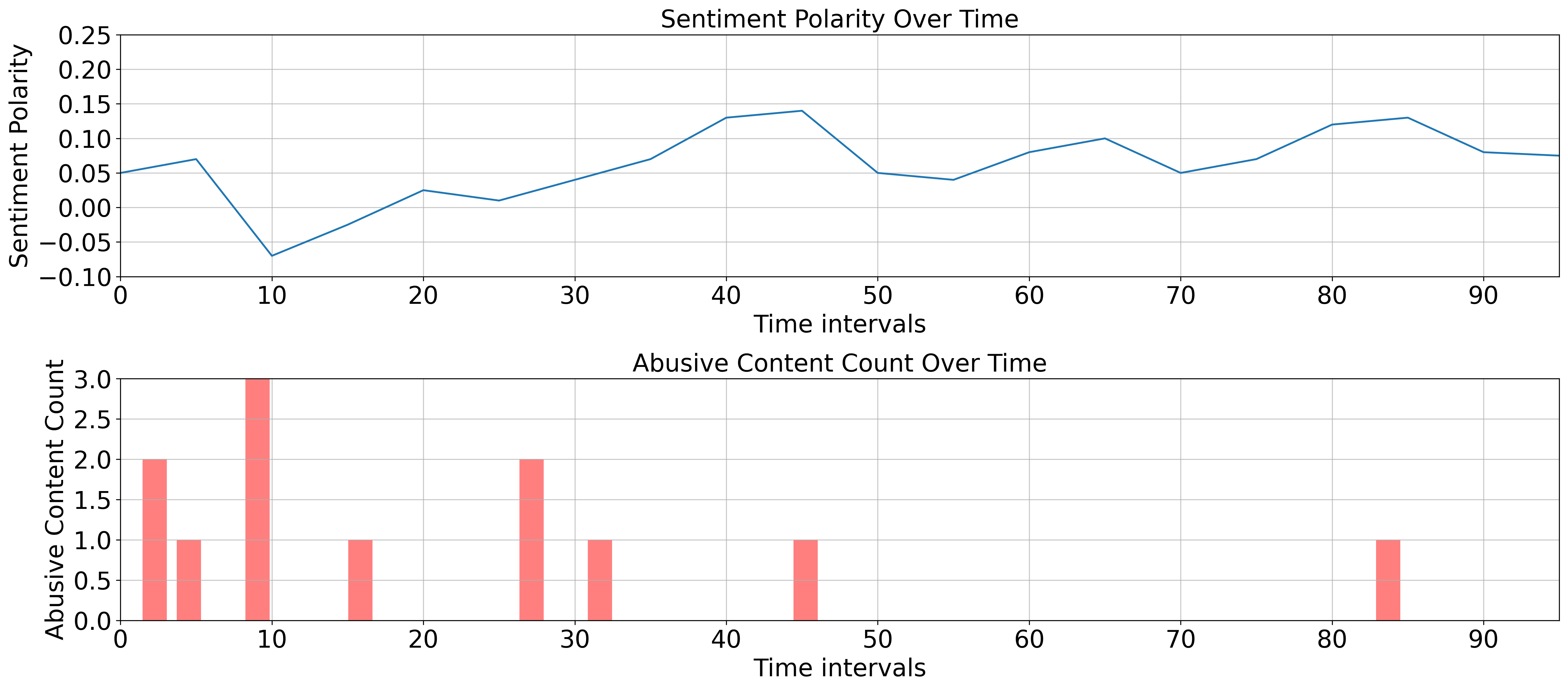}
    \caption{Take "chariots of fire" as an example to conduct sentiment analysis and abusive detection for the entire movie}
    \label{fig:chariots of fire}
\end{figure*}

Figure \ref{fig:The Departed} shows the change in sentiment polarity over time for The Departed, which is an action and drama movie. The sentiment polarity fluctuates considerably throughout the movie's timeline and peaks around the 30th minute. Furthermore, the abusive words are frequent during certain periods of the film, especially between 65-105 minutes and 135-145 minutes. These periods may correspond to high-intensity dialogues or conflict scenes.
A combination of the two graphs suggests a possible correlation between significant decreases in emotional polarity and a high frequency of abusive content or behaviour. 

 Figure \ref{fig:annie hall} presents sentiment polarity and abusive content for the movie Annie Hall. The emotional polarity of the movie fluctuates within a generally positive range, mostly between -0.05 and 0.20. Around the 90th minute, the polarity briefly drops into a negative zone but quickly rebounds, reflecting an emotional low point followed by recovery. This trend aligns with the movie's overall optimistic tone, likely tied to its comedic and lighthearted theme.
The abusive content is concentrated and sporadic, which shows that only a certain part of the film contains relatively concentrated violent or offensive language content. No significant change in sentiment polarity during the period of high incidence of violent content (about 40 to 60 minutes), but there was a small positive fluctuation in emotional polarity during this period.  
The sharp decline in emotional polarity at the 90th minute is not accompanied by the detection of abusive content, suggesting that other non-violent emotional factors, this period may have character conflict and inner struggle in the story.

Figure \ref{fig:the silence of the lambs} presents sentiment polarity and abusive content detection for The Silence of the Lambs. The sentiment polarity undergoes significant fluctuations at several points, with more pronounced positive peaks, especially around periods 30, 65, and 90, suggesting that these moments likely contain more positive emotional content. Abusive content that appears in different periods of the movie. Among them, the time period between 95 and 105 minutes has the most violent content, indicating that this is the film's climax and could be the key violent scene. The violent content also increased during periods 65 and 85, but the intensity of violence was relatively low compared to the peak of 95-105. This may indicate that the film's tension is elevated at these moments, but not at its highest point.

 Figure \ref{fig:chariots of fire} presents sentiment polarity for Chariots of Fire which peaks  at the 240th and 260th minutes capturing moments of heightened emotional intensity to represent emotional climaxes within the film.  
The abusive content detection indicates shows that there is almost no violence in the film, and only a very small amount appears in approximately 150, 215, and 275 minutes. This validates that most of the film revolves around the sportsmanship and inner struggles of the characters, rather than outright violence or conflict.


\section{Discussion}

Our research shows that sentiment polarity analysis underscores the nuanced emotional landscape of films, clearly indicating that movies have become more emotionally complex over the years. The average sentiment scores reveal a trend towards more balanced emotional expressions, with a slight decline in positive sentiment polarity in recent decades. This could be attributed to the increasing depiction of complex and often darker themes in contemporary cinema.
The sentiment co-occurrence analysis revealed intriguing patterns in how emotions interplay within movie dialogues. High co-occurrence frequencies of emotions like joking and annoyance, or sadness and anger, highlight the complex emotional dynamics filmmakers utilise to engage audiences. The rarity of co-occurrences between emotions like thankfulness and pessimism further illustrates the deliberate emotional structuring in cinematic storytelling. 
In related studies, Palomo et al. \citep{Palomo} used deep learning technology to conduct sentiment analysis on IMDb movie reviews and revealed the changes in emotional tendencies in movie reviews of different periods. They reported that from the 1950s to the 1980s, emotional expression underwent significant changes, and since the 1980s, emotional expression has become more complex, especially under the influence of social events such as the global financial crisis, climate change and pandemics.  The impact of negative emotions has become more pronounced, given the rise of films featuring catastrophic natural events, poor governance and politics.

Regarding detection of abuse, our study overlaps with Demarty et al. \citep{Demarty2014} who provided a study on multimodal violence detection by combining video, audio and text subtitle analysis. They reported that the accuracy of violence detection can be significantly improved by integrating visual, audio and language features.  A multimodal detection framework \citep{wu2020not} from 2020 showed that combining audio and visual cues under weak supervision significantly improved violence recognition performance. Information from the different modes complements each other, helping to capture more complex features of violent scenes. However, our approach pays more attention to the combination of text and context. There have been fluctuations in levels of violence in the two major categories (Oscar-nominated and blockbusters), and the level of abuse for Oscar movies took over blockbusters since 2010 (Figure 11). This could be due to over-commercialisation, the rise of social media with mobile phones, and competition with OTT \citep{dhiman2023critical,simlote2024evolution} and other streaming platforms, including Instagram Stories, YouTube Reels, and TikTok. Sharma et al. \citep{sharma2021emerging}   highlighted the rise of abuse and violence in relation to women, caste and religion through Indian web series on OTT platforms. Due to the easy accessibility of OTT platforms, cinema-released movies have been facing challenges in attracting viewers to the cinema, as the movies are often released on OTT platforms shortly after their cinema release. The normalisation of abuse and hate speech, particularly in OTT platforms, could be one of the reasons traditional movies have seen an increase in abusive content over the last decade, and due to their popularity, they have been making it to the Oscar awards.

 Our findings offer several practical implications for stakeholders in the film and media industry. First, film producers and streaming platforms can leverage the observed trends in sentiment polarity and abusive content to refine their content moderation, audience targeting, and storytelling strategies. For example, the increase in abusive language in recent decades highlights the importance of adopting more sophisticated content filtering and age-rating systems on OTT platforms. Second, regulators may consider how the meaning of abusive language evolves when designing classification guidelines, ensuring that content ratings remain relevant to contemporary social norms. Finally, the observed cross-genre and award/commercial differences suggest that marketing strategies should be genre-sensitive, leveraging insights into emotional dynamics to engage audiences more effectively.

Although our study provides a comprehensive overview of the emotional and abusive content in Hollywood films, there are several limitations.   Firstly, the movie data is limited to subtitles of Oscar nominations and the top yearly blockbuster movies. Therefore, they do not cover all types of films and cultural backgrounds. Oscar films tend to have a specific artistic and cultural bent, while the top ten films at the box office may be more commercial and entertaining. As a result, these data may not reflect niche films' emotional and abusive content characteristics. In addition, we only selected films with English subtitles, excluding other languages, contributing to our findings' limitations. Secondly, as a comprehensive art form, the audience is not only affected by dialogue but also by a variety of factors, including vision. We only analysed the film's subtitles, ignoring visual information and description of the scenes where no dialogue takes place. The description of violence in the film is usually conveyed through visual effects, including action scenes, blood effects, and violent action scenes.  Finally, there are limitations regarding the datasets used for fine-tuning language models. The data to fine-tune abusive detection came from messages on the Reddit online community from January 2012 to June 2015, which is not comprehensive. The meaning and use of abusive content have changed significantly over time. Some words may be considered highly insulting in one era, but may have lost their original strong insult in another. In addition, the abusive content dataset used in the study is largely based on contemporary language norms, which may not accurately reflect abusive words and their contexts from different eras in history.

The use of abusive and vulgar language in cinema movies and OTT platforms has become a topic of controversy and debate in recent years \citep{dhiman2023critical}. Unlike traditional television networks, OTT platforms are not subject to the same regulations, allowing them greater freedom in producing and distributing content. Fortunately, most OTT platforms have age ratings and content warnings to help viewers make informed decisions about the content they consume. In response to concerns about using vulgar language on OTT platforms,  government regulators have introduced new rules and guidelines \citep{baccarne2013television,baldry2014rise}. Developing rules and guidelines to regulate content on OTT platforms is an issue that requires urgent attention \citep{dhiman2023critical} with more and more people using OTT platforms. A strict content vetting mechanism needs to be established to ensure that all content uploaded to the platform is vetted. The vetting should include legal compliance, morality and ethics, violence and pornographic content, etc., to safeguard the legality and compliance of the content. 

Another limitation arises from potential biases in both the subtitle corpus and the fine-tuning datasets. Subtitles from OpenSubtitles may differ in accuracy, with potential omissions, simplifications, or translator-specific interpretations, which could affect the fidelity of dialogue analysis. Moreover, our fine-tuning corpora (SenWave and RAL-E) are rooted in English-language social media, limiting the generalizability of findings to other linguistic and cultural settings. As a result, while our study offers meaningful insights into Hollywood English films, caution must be applied when extrapolating results to non-English or culturally diverse contexts. Future research could incorporate multilingual subtitle corpora and culturally diverse datasets to improve robustness and cross-cultural generalizability.

In future research, sentiment analysis and violence detection can use multimodal data analysis \citep{ngiam2011multimodal,tavakoli2023multi}, that is, not only text data but also visual and audio data. More comprehensive emotional information can be obtained by analysing visual and auditory elements (such as expressions, tone of voice, etc.) in movies. In addition, the research can further refine the classification of emotion and can be extended to films in different cultural and regional backgrounds to compare the expression of emotion and the use of abusive content in different cultures. This will help to understand the cultural influence on the film's content. At the same time, developing a real-time analysis system is also a necessary research direction, which can conduct instant emotional and abusive content analysis of newly released movies, and provide immediate review and recommendation to subscribers. Finally, through the knowledge of psychology, sociology and other fields, in-depth research can be done on the impact of film content on audience psychology and behaviour, especially the impact of abusive and violent content on teenagers.

\section{Conclusions}

We employed language models for a longitudinal study of abuse detection and sentiment analysis of subtitle dialogues from prominent Hollywood movies spanning seven decades (1950-2024), including  Oscar-nominated and top ten blockbuster movies. We  refined language models using specific sentiment analysis and abusive datasets to discern significant trends and shifts in the emotional and abusive content within movie dialogues.

Our results reveal a complex landscape of emotional expression in movies, with notable fluctuations in the prevalence of various sentiments and abusive content. Our findings indicate that while positive emotions such as humour and optimism have consistently been integral to movie dialogues.  The representation of negative emotions such as anger, sadness, and anxiety has varied significantly, often reflecting broader social and cultural changes. Our categorisation of films into four broad genres—action, comedy, drama, and thriller—provided further granularity to our analysis. Each genre exhibited distinct patterns of emotional and abusive content, with overlaps, since the majority of movies can be classified into two categories.  

One of the key insights is the gradual increase of abusive content in movie dialogues over time, especially in the last two decades, where Oscar-nominated movies overtook the top ten blockbusters. This trend suggests a gradual shift in the industry's approach to depicting violence and abuse, potentially influenced by evolving societal norms and regulatory frameworks. Our study relies solely on textual subtitles and does not incorporate audio or visual modalities, which are crucial for a more holistic interpretation of sentiment and violence in films. Future work can explore multimodal approaches that integrate speech prosody, facial expressions, and visual context, enabling a more comprehensive analysis of cinematic content.

\section{Code and Data}
The study can be further expanded with the availability of our open-source code and data \footref{github}. The data extracted for this study has been curated in Kaggle \footnote{\url{https://www.kaggle.com/datasets/mlopssss/subtitles}}.

\section{Acknowledgement}

We thank UNSW School of Mathematics and Statistics, DATA-9020 Group-H members: Jiaying Zhu, Yuxin Hou, Ziyang Feng, Zimeng Luo, for their contributions earlier in the project.

 \bibliography{cas-refs,chandra}

\end{document}